\documentclass{article}

\usepackage{microtype}
\usepackage{graphicx}
\usepackage{subfigure}
\usepackage{booktabs} 

\usepackage{hyperref}

\usepackage[accepted]{icml2023}

\usepackage{amsmath}
\usepackage{amssymb}
\usepackage{mathtools}
\usepackage{amsthm}

\usepackage[capitalize,noabbrev]{cleveref}

\theoremstyle{plain}

\theoremstyle{definition}

\theoremstyle{remark}

\usepackage[textsize=tiny]{todonotes}

\usepackage{multirow}
\usepackage{xspace}

\icmltitlerunning{Grounding Language Models to Images for Multimodal Inputs and Outputs}

\def\ModelName{FROMAGe\xspace}

\begin{document}

\twocolumn[
\icmltitle{Grounding Language Models to Images for Multimodal Inputs and Outputs}



\icmlsetsymbol{equal}{*}

\begin{icmlauthorlist}
\icmlauthor{Jing Yu Koh}{cmu}
\icmlauthor{Ruslan Salakhutdinov}{cmu}
\icmlauthor{Daniel Fried}{cmu}
\end{icmlauthorlist}

\icmlaffiliation{cmu}{Carnegie Mellon University}

\icmlcorrespondingauthor{Jing Yu Koh}{jingyuk@cs.cmu.edu}

\icmlkeywords{Machine Learning, Vision-and-Language, Large Language Models}

\vskip 0.7in
]



\printAffiliationsAndNotice{}

\begin{abstract}
We propose an efficient method to ground pretrained text-only language models to the visual domain, enabling them to process arbitrarily interleaved image-and-text data, and generate text interleaved with retrieved images. Our method leverages the abilities of language models learnt from large scale text-only pretraining, such as in-context learning and free-form text generation. 
We keep the language model frozen, and finetune input and output linear layers to enable cross-modality interactions. This allows our model to process arbitrarily interleaved image-and-text inputs, and generate free-form text interleaved with retrieved images. 
We achieve strong zero-shot performance on grounded tasks such as contextual image retrieval and multimodal dialogue, and showcase compelling interactive abilities. 
Our approach works with any off-the-shelf language model and paves the way towards an effective, general solution for leveraging pretrained language models in visually grounded settings.
\end{abstract}

\section{Introduction}

Trained at massive scale on large text corpora, large language models (LLMs) are able to demonstrate compelling abilities such as generating human-like dialogue and answering complex questions. While undeniably impressive, most state-of-the-art LLMs are trained on text-only data scraped from the Internet. They are not exposed to rich visual cues, and are often unable to learn concepts \textit{grounded} in the real world. Consequently, most existing language models exhibit limitations on tasks that involve visual reasoning and grounding, and they are also incapable of producing images.

\begin{figure}
    \centering
    \includegraphics[width=1.0\linewidth]{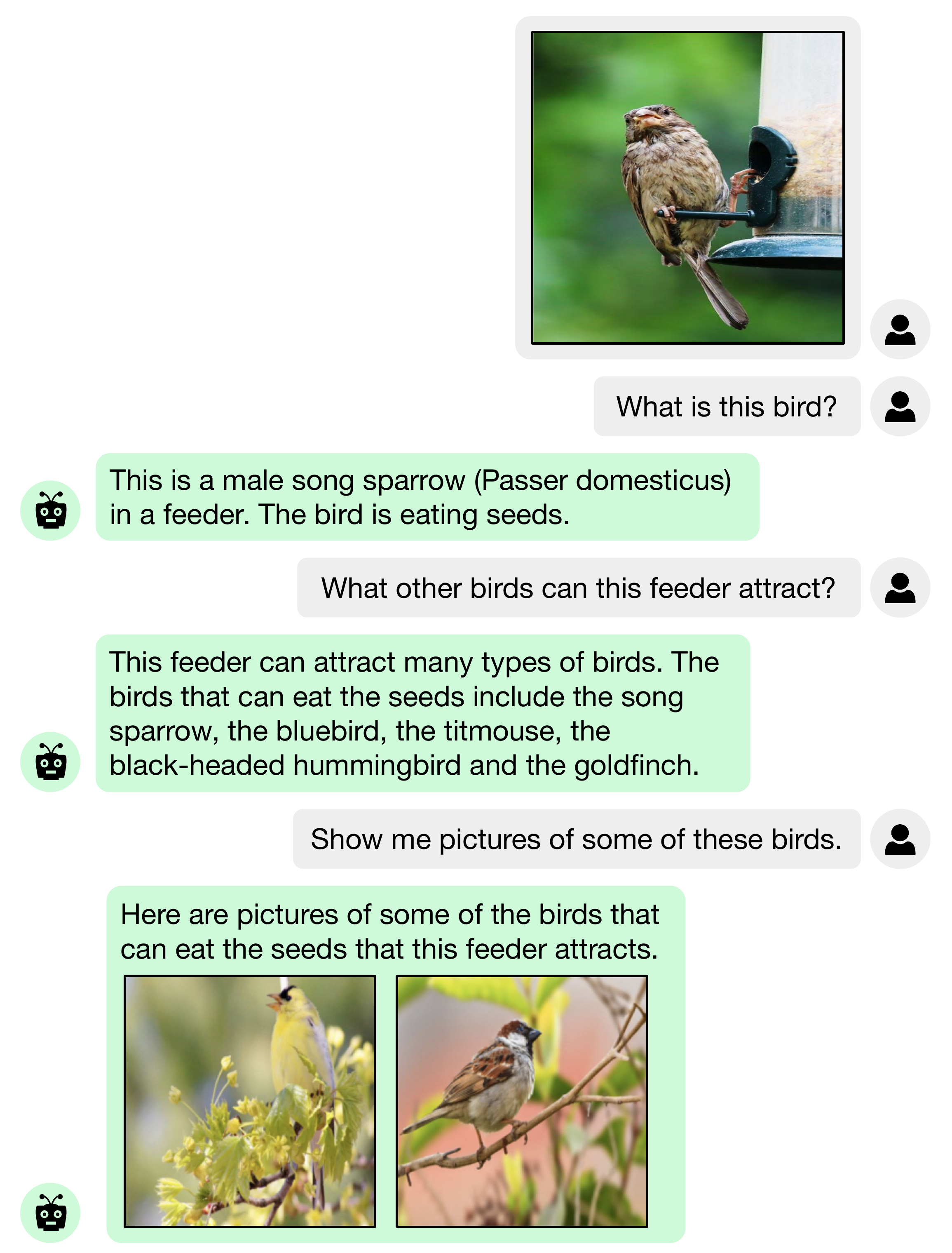}
    \vspace{-0.3in}
    \caption{Our method grounds a language model to the visual domain, enabling it to process arbitrarily interleaved image-text inputs and generate coherent text outputs interleaved with relevant retrieved images. Speech bubbles in green are model generated, while grey bubbles are input prompts.}
    \vspace{-0.15in}
    \label{fig:figure1}
\end{figure}

In this paper, we show that we are able to efficiently leverage the capabilities of a frozen LLM for multimodal (image and text) input and output. 
Our approach equips text-only models with exciting new vision-and-language capabilities such as multimodal dialogue, generation, and contextual image retrieval from conversations (Fig.~\ref{fig:figure1} and Fig.~\ref{fig:qualitative_results}).

We propose a method to bootstrap a frozen language model for processing and outputting arbitrarily interleaved multimodal data. 
We start from a frozen pretrained LLM, and a frozen pretrained visual encoder, and train with a multi-task objective for (1) image captioning (learning to process interleaved multimodal inputs) and (2) image-text retrieval (learning to produce interleaved multimodal outputs). 
For captioning, we extract visual embeddings from the visual encoder, and learn a linear mapping through the maximum-likelihood objective to map embeddings into the input space of the language model. For image-text retrieval, we train the language model to learn a new \texttt{[RET]} token which represents an image, and learn a linear mapping through contrastive learning~\cite{oord2018representation} to map the \texttt{[RET]} embeddings for a caption to be close to the visual embeddings for its paired image. 
Most of the model is kept frozen, and we only update the weights of the linear layers and the \texttt{[RET]} token embedding during training. Hence, our proposed method is very computationally and memory efficient.\footnote{Our model is trained in less than 24 hours on a single GPU.} Once trained, our model exhibits several capabilities. It retains the original abilities of the text-only LLM to generate text, but also attains new multimodal dialogue and reasoning abilities. Our proposed method is model agnostic, and can be applied to ground larger or stronger LLMs released in the future. 
Our main contributions include:
\begin{itemize}
    \item Proposing \textbf{F}rozen \textbf{R}etrieval \textbf{O}ver \textbf{M}ultimodal Data for \textbf{A}utoregressive \textbf{Ge}neration (\ModelName), a model efficiently trained by visually grounding LLMs with image captioning and contrastive learning.
    \ModelName learns strong few-shot multimodal abilities from image-caption pairs alone, while other models require web-scale interleaved image-text data~\cite{alayrac2022flamingo,aghajanyan2022cm3}.
    \item Demonstrating that autoregressive LLMs can perform text-to-image retrieval with greater sensitivity to input text. Our approach is more accurate on long and complex free-form text compared to existing models.
    \item Showing that the existing capabilities of pretrained text-only LLMs, such as in-context learning, input sensitivity, and dialogue generation, can be leveraged for visually grounded tasks. 
    We demonstrate: (1) contextual image retrieval given sequences of interleaved images and text, (2) strong zero-shot performance on visual dialogue, and (3) improved sensitivity to discourse context for image retrieval.
\end{itemize}

Our findings pave the way towards models capable of conditioning on and generating long, coherent, multimodal sequences, and provide further insights into the abilities of pretrained text-only LLMs on visually grounded tasks. Our code and pretrained models are made publicly available\footnote{\url{https://github.com/kohjingyu/fromage}} to encourage future work and exploration.

\section{Related Work}

\paragraph{Large language models.}
Large language models have recently received significant attention in the machine learning and natural language processing communities, in part due to their intriguing abilities to perform in-context learning~\cite{brown2020language,chan2022data} and long-form generation~\cite{dai2019transformer,tan2020progressive,yang2022re3}. Most state-of-the-art models are variants of the Transformer model~\cite{vaswani2017attention}, with the best models achieving gains from scaling model size~\cite{rae2021scaling,smith2022using,chowdhery2022palm}, increasing pretraining data~\cite{hoffmann2022training}, improving finetuning objectives~\cite{wei2021finetuned,tay2022transcending}, and more.

\paragraph{LLMs for vision-and-language.} The strong performance of LLMs has also inspired work in vision-and-language research. 
DALL-E~\cite{ramesh2021zero} proposed a Transformer based model for text-to-image generation~\cite{reed2016generative} by treating images as sequences of discrete tokens. This framework was improved upon by other methods~\cite{yu2022scaling,ding2022cogview2} through model scaling, pretraining, and improved image quantization models~\cite{esser2021taming,yu2021vector}. Flamingo~\cite{alayrac2022flamingo} proposed a visual language model for text generation, with an impressive ability to adapt to and achieve state-of-the-art on a variety of vision-and-language tasks. Several other approaches~\cite{wang2022unifying,li2022blip} also propose multi-task vision-and-language pretraining approaches to improve model performance.  CM3~\cite{aghajanyan2022cm3} trained a causally masked model on a large HTML corpus, and showed that the model is capable of generating images and text.
We differ from previous work in that our model is capable of generating coherent multimodal outputs: Flamingo is incapable of producing visual outputs, while CM3 generally produces poor visual outputs (further comparison in the appendix). In addition, \ModelName is efficient and requires significantly less compute: it is trained in 1 GPU day (Flamingo uses 1535 TPUs for 15 days, and CM3 uses 384 GPUs for 24 days), and does not require web-scale interleaved image-text data. 

\paragraph{Efficient adaptation of pretrained models.} Lastly, our work builds upon approaches for parameter and resource efficient adaptation of pretrained models. Prefix and prompt tuning~\cite{lester2021power,li2021prefix} enable adaptation of a pretrained LLM to new settings by finetuning a small set of parameters to act as an input prefix, while keeping the rest of the model parameters frozen. \citet{houlsby2019parameter} proposed adapters for transferring pretrained LLMs to new language tasks. Frozen~\cite{tsimpoukelli2021multimodal} proposed training a visual encoder to enable few-shot learning for multimodal tasks. 
MAGMA~\cite{eichenberg2021magma} improved upon Frozen by training adapter modules for improved performance on downstream tasks. 
ESPER~\cite{yu2022multimodal} uses reinforcement learning to improve zero-shot transfer and caption style transfer. 
\citet{lu2021pretrained} show that language-pretrained transformers can transfer well to non-language tasks. 
LIMBeR~\cite{merullo2022linearly} analyzes pretrained vision and language models, and finds that learnt representations are functionally equivalent up to a linear transform. 
Our work builds upon the insights and methods from these prior works. While previous models mostly focus on generating text-only outputs, our model is capable of processing arbitrarily interleaved image-text inputs to generate coherent interleaved image-text outputs.

\section{Method}
\begin{figure*}[t]
    \centering
    \includegraphics[width=1.0\textwidth]{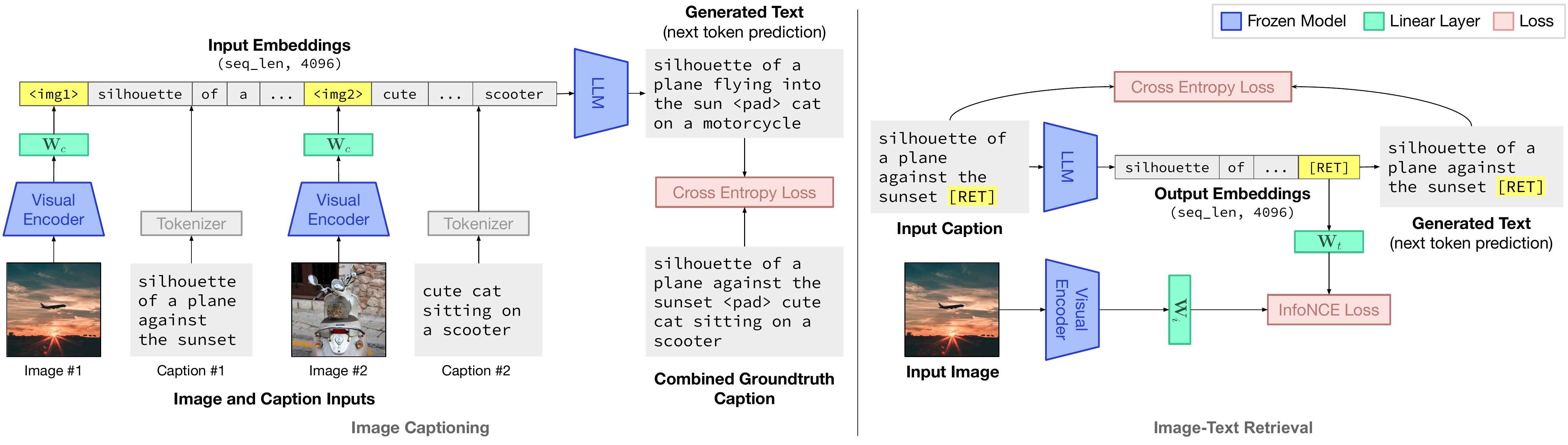}
    \vspace{-0.3in}
    \caption{Overview of the \ModelName architecture. \ModelName is a model trained on image-text pairs for image captioning and image-text retrieval. It is capable of processing arbitrarily interleaved image and text inputs, and producing interleaved images and text as outputs.}
    \vspace{-0.1in}
    \label{fig:architecture}
\end{figure*}

\begin{figure*}
    \centering
    \includegraphics[height=0.98\textheight]{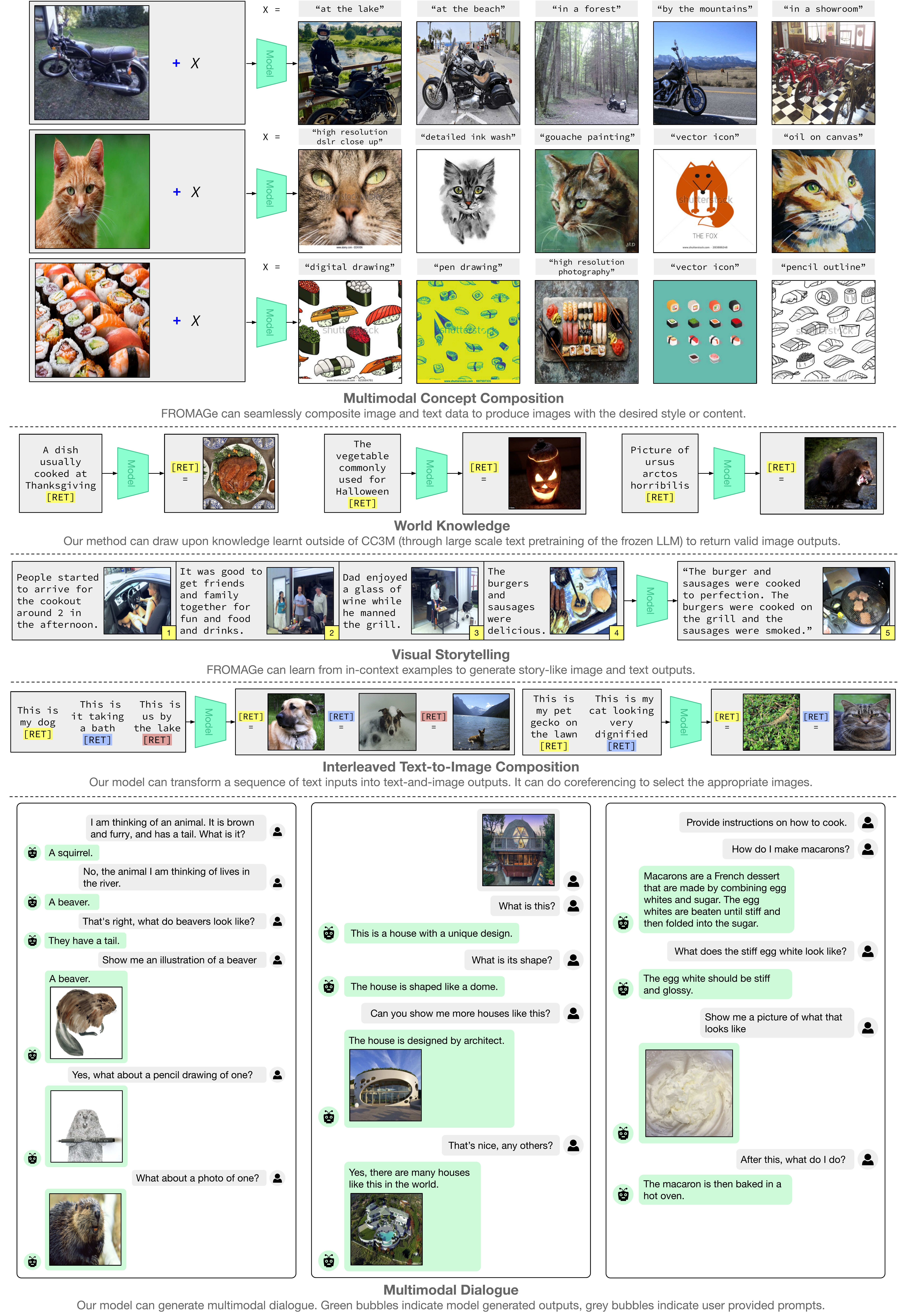}
    \vspace{-0.15in}
    \caption{Selected examples from \ModelName for various image-text tasks. \ModelName is sensitive to context: it can generate multimodal dialogue, and rapidly learn in-context to perform various few shot image-text tasks. More examples are provided in the appendix.}
    \label{fig:qualitative_results}
\end{figure*}

Our approach integrates a language model and visual model while keeping their parameters frozen. We learn translation parameters (parameterized as linear layers) to cast images into text space, and text embeddings into visual space. Our motivation for keeping the models frozen is to leverage the capabilities of the LLM learnt from large scale pretraining. 
We find that this enables better generalization to zero-shot and few-shot settings (further analysis in Sec.~\ref{sec:ablation_analysis}). 

\subsection{Model Architecture}

\paragraph{Language model.} \ModelName takes an autoregressive large language model $p_{\theta}$, originally trained with the max-likelihood objective on text-only data, and keeps its parameters $\theta$ frozen. Given text $x$ (e.g., an image caption), the models we use extract a sequence of input tokens $(s_{1}, \ldots, s_{T})$ using a byte-level BPE tokenizer~\cite{sennrich2015neural,radford2019language,brown2020language}. 
The models were trained to maximize the log likelihood of the token sequence, factorized as a sum of conditional log probabilities:
\begin{align*}
\log p_{\theta}(x) = \sum_{t=1}^t \log p_{\theta}(s_{t} | s_{1}, \ldots, s_{t-1})
\end{align*}
\paragraph{Visual model.} To extract visual information from an input image $y$ corresponding to a caption $x$, we use a pretrained visual backbone model which produces visual embeddings $v_{\phi}(y) \in \mathbb{R}^m$. The weights $\phi$ are kept frozen as well.

\subsection{Translating Between Image-and-Text} \label{sec:translation_params}

To integrate vision and language, we learn translation parameters to map between the image and text embedding spaces. This extends a LLM for multimodal inputs and outputs.

\paragraph{Mapping image-to-text.} 
We learn a linear mapping $\mathbf{W}_c \in \mathbb{R}^{m \times kd}$ which maps visual embeddings $v_{\phi}(y)$ from the visual model for image $y$ as $v_{\phi}(y)^T \mathbf{W}_c \in \mathbb{R}^{k \times d}$ (after reshaping $kd$ to $k \times d$). This represents a sequence of $k$ vectors of the same hidden dimensionality $d$ as the text embeddings that the LLM produces for input tokens.

\paragraph{Mapping text-to-image.} Our approach aims to retrieve images using the outputs of an autoregressive language model. A challenge with this is that autoregressive causal attention over text is strictly less expressive than the bidirectional attention typically used in previous models~\cite{radford2021learning,jia2021scaling}. In order to bootstrap strong retrieval abilities on our autoregressive LLM, we propose adding a special \texttt{[RET]} token to the model vocabulary and learning its embeddings (keeping all other token embeddings frozen). During training, we append \texttt{[RET]} to the end of input captions. This allows the model to perform an extra step of attention over all tokens in the text to produce a stronger text representation for the caption. We found that this significantly improves image-text retrieval performance (see Sec.~\ref{sec:ablation_analysis} for analysis). This also allows our model to learn to generate \texttt{[RET]} at inference time (Fig.~\ref{fig:figure1}), seamlessly interleaving image retrieval within generated text.

Finally, to map the model's output representations to visual space, we train a linear mapping $\mathbf{W}_t \in \mathbb{R}^{p \times q}$. This maps the hidden representation of \texttt{[RET]} from the last hidden layer of the LLM, $h_{\theta}(x_i) \in \mathbb{R}^{p}$, into a vector space for retrieval, where $q$ is a dimension smaller than $p$. Similarly, we train another linear mapping $\mathbf{W}_i \in \mathbb{R}^{m \times q}$ to map the visual embeddings $v_{\phi}(y_i)$ into the same retrieval space.

\subsection{Training Setup}  \label{sec:training_setup}
We train \ModelName with a multi-task objective of image captioning and image-text retrieval (summarized in Fig.~\ref{fig:architecture}):

\paragraph{Image captioning.} Similar to previous work~\cite{tsimpoukelli2021multimodal,eichenberg2021magma}, we formulate image captioning as the task of generating text tokens conditioned on a visual prefix. The visual prefix is the output of the image-to-text mapping layer, $v_{\phi}(y)^T \mathbf{W}_c$, which is prepended to the caption. The log likelihood of caption $x$ (tokenized as $(s_{1}, \ldots, s_{T})$) conditioned on its image $y$ is:
\begin{align*} 
l_{c}(x, y) = \sum_{t=1}^T \log p_{\theta}(s_{t} | v_{\phi}(y)^T \mathbf{W}_c, \; s_{1}, \ldots, s_{t-1})
\end{align*}
The captioning loss $\mathcal{L}_{c}$ is then the negative log likelihood of all samples in a batch of $N$ image-text pairs:
\begin{align} \label{eq:captioning_loss}
\mathcal{L}_{c} = - \frac{1}{N} \sum_{i=1}^N l_c(x_i, y_i)
\end{align}

\paragraph{Image-text retrieval.} In addition to image captioning, we train our model to retrieve images conditioned on text (and vice versa). Image-text retrieval has been used to learn joint visual and language representations~\cite{jia2021scaling,radford2021learning}, enabling cross-modality search from text descriptions. Underlying the approach is contrastive learning~\cite{chopra2005learning} with the InfoNCE loss~\cite{oord2018representation}. 
Given a caption $x$ and its paired image $y$, we extract the output of the last hidden layer of the LLM for the \texttt{[RET]} token, $h_{\theta}(x)$, and the output of the visual encoder for the image, $v_{\phi}(y)$. The normalized cosine similarity for the image and text embeddings can be computed with the learnt linear mappings $\mathbf{W}_t$, and $\mathbf{W}_i$ (described in Sec.~\ref{sec:translation_params}):
\begin{align*}
\text{sim}(x, y) = \frac{(h_{\theta}(x)^T \mathbf{W}_t) (v_{\phi}(y)^T \mathbf{W}_i)^T}{ \lVert h_{\theta}(x)^T \mathbf{W}_t \rVert \lVert v_{\phi}(y)^T \mathbf{W}_i)^T \rVert }
\end{align*}
We minimize the InfoNCE loss for text-to-image (t2i) and image-to-text (i2t) retrieval over a batch of $N$ text-image pairs $(x_i, y_i)$, where each example is treated as a positive pair, and other in-batch examples as negatives:
\begin{align}  
\mathcal{L}_{\text{t2i}} &= -\frac{1}{N} \sum_{i=1}^N \left( \log \frac{\exp(\text{sim}(x_i, y_i) / \tau)}{ \sum_{j=1}^N \exp(\text{sim}(x_i, y_j) / \tau )} \right)  \label{eq:t2i_loss} \\
\mathcal{L}_{\text{i2t}} &= -\frac{1}{N} \sum_{i=1}^N \left( \log \frac{\exp(\text{sim}(y_i, x_i) / \tau)}{ \sum_{j=1}^N \exp(\text{sim}(y_i, x_j) / \tau )} \right)  \label{eq:i2t_loss}
\end{align}

Similar to previous work~\cite{jia2021scaling,radford2021learning}, $\tau$ is a learnable temperature parameter.  The final training loss is a weighted sum of the captioning loss (Eq.~\ref{eq:captioning_loss}) and the retrieval losses (Eq.~\ref{eq:t2i_loss} and \ref{eq:i2t_loss}):
\[ \mathcal{L} = \lambda_c \mathcal{L}_{\text{c}} + \lambda_r (\mathcal{L}_{\text{t2i}} + \mathcal{L}_{\text{i2t}})  \]

$\lambda_c$ and $\lambda_r$ are hyperparameters representing captioning and retrieval loss weights respectively. Since $\theta$ and $\phi$ are frozen, only the linear mappings $\mathbf{W}_c$, $\mathbf{W}_t$, and $\mathbf{W}_i$, and the \texttt{[RET]} embedding vector receive gradient updates.

\subsection{Data and Implementation Details}
We train on the Conceptual Captions (CC3M) dataset~\cite{sharma2018conceptual} consisting of 3.3 million image-text pairs.\footnote{3.1M examples remain after filtering out missing images.} To encourage the model to attend more explicitly to images, we randomly concatenate distinct examples together (with probability of 0.5 to concatenate) for the image captioning task (Fig.~\ref{fig:architecture}, left). We found this helpful in training the model to attend to the correct image within a sequence (detailed analysis in the appendix).

We use the publicly available OPT model~\cite{zhang2022opt} with 6.7B parameters as our LLM. Past work indicates that findings at the 6.7B scale are likely to generalize to larger model sizes~\cite{dettmers2022llm}, and large enough to exhibit the few-shot and in-context learning abilities that we are interested in~\cite{radford2019language}. For the visual model, we use a pretrained CLIP ViT-L/14 model~\cite{radford2021learning} for its ability to produce strong visual representations for vision-and-language tasks~\cite{merullo2022linearly}.

All models are implemented in PyTorch~\cite{paszke2019pytorch} v1.12 and trained mixed-precision with bfloat16~\cite{abadi2016tensorflow}. As most of the model parameters (97.0\%) are frozen, our method is memory and compute efficient: we backpropagate through the frozen LLM and visual model, but only compute gradient updates for the 3 trainable linear mappings and \texttt{[RET]} embedding (see Sec.~\ref{sec:training_setup}). Our models are trained with a batch size of 180 for 1 epoch (18000 iterations) on 1 A6000 GPU (clock time of 24 hours). We use the Adam~\cite{kingma2014adam} optimizer with a learning rate of $0.0003$ and warmup of 100 steps. The loss weights $\lambda_c$ and $\lambda_r$ are set to 1 and we use a visual prefix length of $k=1$ and retrieval embedding dimension $q = 256$, and embedding dimension $d = 4096$ (inherited from OPT-6.7B).

\section{Experiments}

The most interesting capabilities of \ModelName emerge in situations with both image-and-text inputs and image-and-text outputs, such as multimodal dialogue (Fig.~\ref{fig:figure1}). As there does not exist comprehensive benchmarks for these specific tasks, we focus evaluation on image retrieval and image-and-text generation tasks, as few prior models are capable of this. 
We benchmark performance on various configurations of multimodal inputs, detailed in the following sections.

\subsection{Contextual Retrieval from Multimodal Inputs}  \label{sec:contextual_image_retrieval}
\begin{figure}
    \centering
    \includegraphics[width=1.0\linewidth]{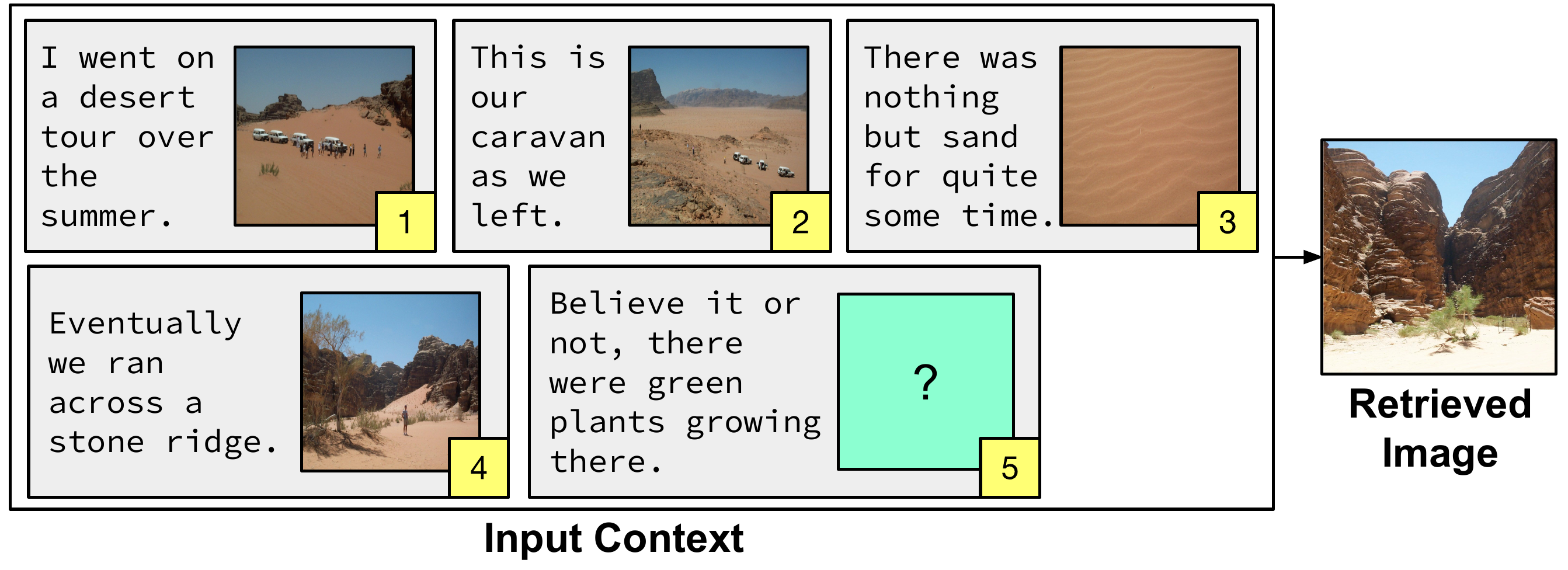}
    \vspace{-0.25in}
    \caption{Contextual image retrieval conditioned on a Visual Story~\cite{huang2016visual} of interleaved image-and-text inputs.}
    \vspace{-0.1in}
    \label{fig:vist_retrieval}
\end{figure}  

Prior work on image-text retrieval~\cite{radford2021learning,jia2021scaling} typically focuses on retrieving a single image from a single caption (and vice versa). \ModelName is adapted from a frozen LLM, and we find that it inherits several interesting behaviors of LLMs, such as in-context learning and greater sensitivity to input context. This benefits many downstream applications, such as multimodal dialogue or image-and-text generation~(examples in Fig.~\ref{fig:qualitative_results}).

In order to evaluate the abilities of \ModelName to process multimodal contextual information, we assess its performance in retrieving the appropriate image conditioned on a sequence of interleaved image-text inputs from the Visual Storytelling (VIST) dataset~\cite{huang2016visual}. Each example in VIST consists of 5 images and text pairs in temporal order (Fig.~\ref{fig:vist_retrieval}). VIST examples are of ``stories'', which are of a very different style compared to the image caption data \ModelName is trained on. This allows us to evaluate our model's capability for in-context learning and zero-shot transfer. This also acts as an evaluation for discourse capabilities, as VIST contains more free-form text. We benchmark over several different experimental setups:
\begin{enumerate}
    \setlength{\itemsep}{0pt}
    \item Retrieve the last image correctly given \textbf{its description}. This is similar to standard image-text retrieval. 
    \item Retrieve the last image given \textbf{the 5 preceding descriptions.} Image-text pairs in VIST follow a temporal order. This tests the ability of retrieval models to condition on free-form temporally dependent language.
    \item  Retrieve the last image given \textbf{the 5 preceding descriptions and 4 images.} This tests the ability of retrieval models to process interleaved image-and-text context.
\end{enumerate}

\addtocounter{footnote}{-1}
\begin{table}[t]
\begin{center}
\setlength{\tabcolsep}{3pt}
\scriptsize
\resizebox{1.0\linewidth}{!}{%
\begin{tabular}{lcccc}
\noalign{\smallskip}
\toprule
\textbf{Model} & \textbf{Inputs}  &  \textbf{R@1} & \textbf{R@5} &  \textbf{R@10} \\ \midrule
CLIP ViT-L/14  &  \multirow{2}{*}{1 caption}   &  \textbf{11.9} &  \textbf{25.5} & \textbf{32.2}  \\
\ModelName  &  &   11.3   &  24.6  &   32.1 \\   \midrule  
CLIP ViT-L/14  &  \multirow{2}{*}{5 captions}   &  5.9 &  19.5 & 28.0  \\
\ModelName  &   &  \textbf{11.9}   &  \textbf{23.8}  &   \textbf{31.7}  \\ \midrule
BLIP$^{\dagger}$  &  5 captions  &  6.2  &  16.8 &  23.4 \\  
CLIP ViT-L/14$^{\dagger}$  &  5 captions  &  8.8  &  22.3 &  29.8 \\  
\ModelName$^{\dagger}$  &  5 captions  &   13.2  &  28.5  &   36.7  \\  
CLIP ViT-L/14  &  5 captions, 4 images  &  2.4   &  21.3   &  34.0   \\
\ModelName$^{\dagger}$ \hspace{5mm}  &  5 captions, 4 images  &   \textbf{18.2}   &  \textbf{42.7}  &   \textbf{51.8}  \\  
\bottomrule
\end{tabular}
}
\vspace{-0.1in}
\caption{Recall@$k$ on zero-shot contextual image retrieval of the last image in Visual Storytelling~\cite{huang2016visual}. Numbers in \textbf{bold} indicate best scores for a particular set of inputs. $^{\dagger}$ indicates retrieval over images not previously seen in the story sequence. \footnotemark}
\label{table:vist_retrieval_results}
\vspace{-0.25in}
\end{center}
\end{table}
\footnotetext{Previous versions of the paper had slightly worse scores due to a normalization bug.}

Our results are presented in Table~\ref{table:vist_retrieval_results}. We primarily compare against CLIP~\cite{radford2021learning}, as it is one of the strongest open sourced and open domain image-text retrieval models available. We report Recall@$k$ (R@$k$) metrics in Tab.~\ref{table:vist_retrieval_results}. 
For a single caption input, CLIP outperforms our model, which we attribute to the CLIP text encoder being a bidirectional model trained specifically for image-text retrieval,\footnote{ For these same reasons, CLIP is unsuitable for dialogue, and does not have few-shot and in-context learning abilities.} while our language model was trained on free-form text. However, as greater context is provided to both models, \ModelName substantially outperforms CLIP. Given the full set of 5 captions, CLIP performance substantially deteriorates (as it appears to be unable to properly handle longer, temporally dependent sentences), with R@1 decreasing by 50.4\% relative to the single caption setting. In contrast, \ModelName is able use the additional descriptions to improve in retrieval accuracy (11.3 to 11.9 on R@1).

\ModelName also effectively conditions on multimodal context (which previous image-text retrieval models are not explicitly trained for). When both images and text inputs are provided to the model, retrieval improves significantly, increasing by 37.9\% on R@1 relative to the caption-only setting  (13.2 to 18.2). Similar improvements are seen on R@5 and R@10. This is a substantial gain over the baseline CLIP model with 5 captions: we achieve a relative improvement of 107\% on R@1 (8.8 to 18.2) when image-and-text context is provided. 

We also run an experiment to test the ability of CLIP to retrieve images conditioned on multimodal context. We embed each of the images and descriptions in the input, and average their embeddings. We find that it does substantially worse than when only caption inputs are provided: it achieves a R@1 of 2.4, a significant decrease from the CLIP R@1 of 8.8 when it is provided with 5 captions. likely because it is trained to condition on image-text inputs. These results are significantly worse than that of \ModelName under the same settings.

These results showcase the efficacy of \ModelName as an image-text model sensitive to complex language descriptions and multimodal context (more analysis in Sec.~\ref{sec:multimodal_context}). It is capable of parsing interleaved multimodal inputs, and strongly outperforms CLIP for longer input descriptions. As for free-form text generation, we also run human evaluations to evaluate the ability of \ModelName to generate stories by learning in-context from VIST examples (Sec.~\ref{sec:multimodal_context}).

\begin{table*}[t]
\begin{center}
\setlength{\tabcolsep}{3pt}
\scriptsize
\resizebox{1.0\linewidth}{!}{%
\begin{tabular}{lcccccccccccccc}
\noalign{\smallskip}
\toprule
 &  &  & &  &  & \multicolumn{5}{c}{\textbf{IT2T}}  & &    \multicolumn{3}{c}{\textbf{T2I}}   \\
\cmidrule{7-11}  \cmidrule{13-15}
\textbf{Model} & & \textbf{Trainable Params} & & \textbf{Finetuning Data} & &   \textbf{NDCG}  & \textbf{MRR} &  \textbf{R@1} & \textbf{R@5} &  \textbf{R@10} & & \textbf{R@1} & \textbf{R@5} &  \textbf{R@10}   \\ \midrule
ViLBERT~\cite{lu2019vilbert}  & & 114M & & 3.1M & &  11.6 & 6.9  &  2.6  &  7.2  &   11.3   & &  -  &  -  &   - \\
CLIP ViT-L/14~\cite{radford2021learning} & & 300M & & 400M & &  10.9 & 8.5  &  3.1  &  8.7  &   15.9   &  & 17.7  &  38.9  &   50.2   \\
Flamingo~\cite{alayrac2022flamingo} & & 10.2B  & & 1.8B  & &  \textbf{52.0} &  -  &  -  &  -  &   -   & & \multicolumn{3}{c}{Incapable}    \\
ESPER~\cite{yu2022multimodal}  & & 4M  & & 0.5M & &  22.3 & \textbf{25.7}  &  14.6  &  -  &   -   & & \multicolumn{3}{c}{Incapable}    \\
\ModelName (ours)  & & 5.5M & & 3.1M & &  16.5  &  22.0 & \textbf{17.6}  &  \textbf{20.1}  &   \textbf{25.1}   & & \textbf{20.8}  &  \textbf{44.9}  &   \textbf{56.0}    \\  
\bottomrule
\end{tabular}
}
\vspace{-0.15in}
\caption{Zero-shot results on Visual Dialog~\cite{das2017visual}, for image-and-text-to-text (IT2T) and text-to-image (T2I) retrieval. 
Unlike previous methods, \ModelName is capable of generating free-form text interleaved with image outputs through text-to-image retrieval.}
\label{table:visdial_results}
\end{center}
\end{table*}

\subsection{Visual Dialogue}

We evaluate \ModelName on zero-shot Visual Dialog (VisDial)~\cite{das2017visual}. We test its ability to (1) select the correct text answer (from 100 candidates) for a question given an image and a conversation about it (image-and-text-to-text, IT2T, which is the standard VisDial task), and (2) retrieve the correct image given a conversation about it (text-to-image, T2I). Our results are summarized in Tab.~\ref{table:visdial_results}. 

For IT2T, since \ModelName is an autoregressive generative model, we compute the perplexity of each question and answer sequence, and select the option with the lowest perplexity. \ModelName outperforms ESPER~\cite{yu2022multimodal}, CLIP~\cite{radford2021learning}, and ViLBERT~\cite{lu2019vilbert} on R@1, improving by 20.5\% relative to ESPER. \ModelName also achieves a competitive Mean Reciprocal Recall (MRR) of 22.0 and Normalized Discounted Cumulative Gain (NDCG) of 16.5. This is substantially higher than ViLBERT and CLIP, but worse than ESPER. We hypothesize that this is due to differences in training: ESPER uses reinforcement learning and trains on MS-COCO (from which VisDial images are derived). 
Flamingo~\cite{alayrac2022flamingo} is substantially better than all other zero-shot models, which we attribute to its larger model size (80B parameters, of which 10.2B are trainable), and larger training data of multimodal webpages (43M webpages) and image-and-text data (1.8B pairs). In contrast, \ModelName has 5M trainable parameters and is trained on CC3M (3.1M image-text pairs). Our approach may also be applied to the Flamingo model (which uses a 70B language model backbone) to enable image retrieval, which is likely to improve overall capabilities and extend it to a greater variety of tasks. 
On the T2I retrieval task, \ModelName significantly outperforms prior work, achieving a 17.5\% relative improvement over CLIP on R@1. ESPER and Flamingo are trained to generate text-only outputs, and are hence incapable of this task.

Our experiments demonstrate that \ModelName achieves competitive results on zero-shot Visual Dialogue. We emphasize that unlike previous models, \ModelName can output \textit{interleaved} image-and-text content, and is a more general model.

\subsection{Qualitative Results}
We also share selected examples covering various interaction settings in Fig.~\ref{fig:qualitative_results}. \ModelName is capable of learning in-context to perform many different zero-shot and few-shot tasks. Many of the most interesting settings are those which produce interleaved images and texts as outputs, which prior work~\cite{tsimpoukelli2021multimodal,alayrac2022flamingo} is incapable of, or does not generate semantically meaningful outputs for (see appendix for further comparison with CM3~\cite{aghajanyan2022cm3}). Our model is capable of holding multimodal dialogue conversations --- processing input images and text and responding with coherent text and image outputs. It is able to refine input images by compositing images and text concepts. \ModelName also inherits the world knowledge of the frozen LLM, and can answer questions that require specific real world facts.

\section{Analysis} \label{sec:analysis}

We analyze various aspects of \ModelName to determine their effects on its overall capabilities. In all experiments, models were trained on CC3M for 24 hours on a single A6000 GPU.

\subsection{Ablation Experiments}  \label{sec:ablation_analysis}

We perform several ablation experiments to validate the design choices made in \ModelName. We provided further details and results of various other ablations in the appendix.

\paragraph{Freezing the LLM.} We find that freezing the language model is essential to retaining in-context learning and few-shot generalization abilities. When finetuned, \ModelName performs significantly worse on VIST and VisDial. Finetuning decreases retrieval performance on VIST (R@1 with full multimodal context decreases from 12.8 to 6.2) as well as VisDial text retrieval (R@1 from 14.6 to 1.0).

\paragraph{Learning a dedicated retrieval token.} As described in Sec.~\ref{sec:translation_params}, we add a special \texttt{[RET]} token to represent embeddings for retrieving images from text. 
When the model is trained without the \texttt{[RET]} token, R@1 performance on VIST (with full multimodal context) is significant worse. We observe that adding the \texttt{[RET]} token improves R@1 by a substantial 38.1\% relative gain over the model without the \texttt{[RET]} token. We observe similar improvements across the board for other tasks.

\subsection{The Effect of Context}  \label{sec:multimodal_context}
\begin{figure}[t]
    \centering
    \includegraphics[width=1.0\linewidth]{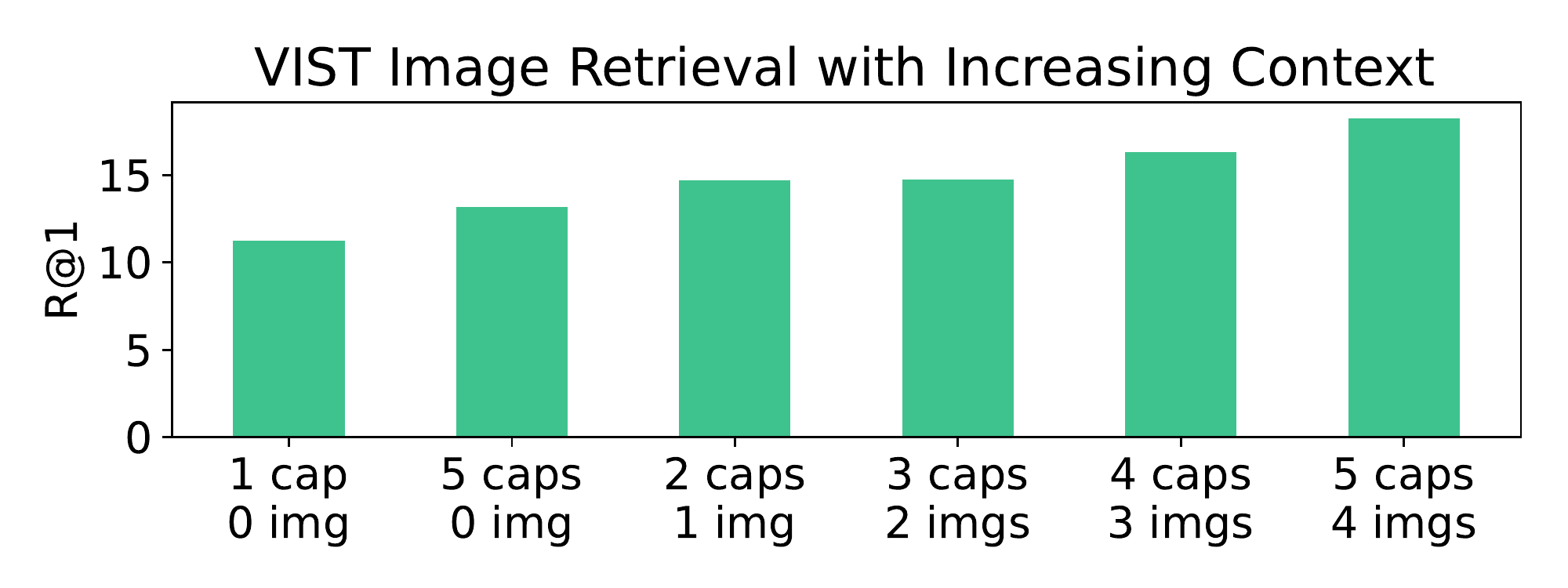}\\
    \includegraphics[width=1.0\linewidth]{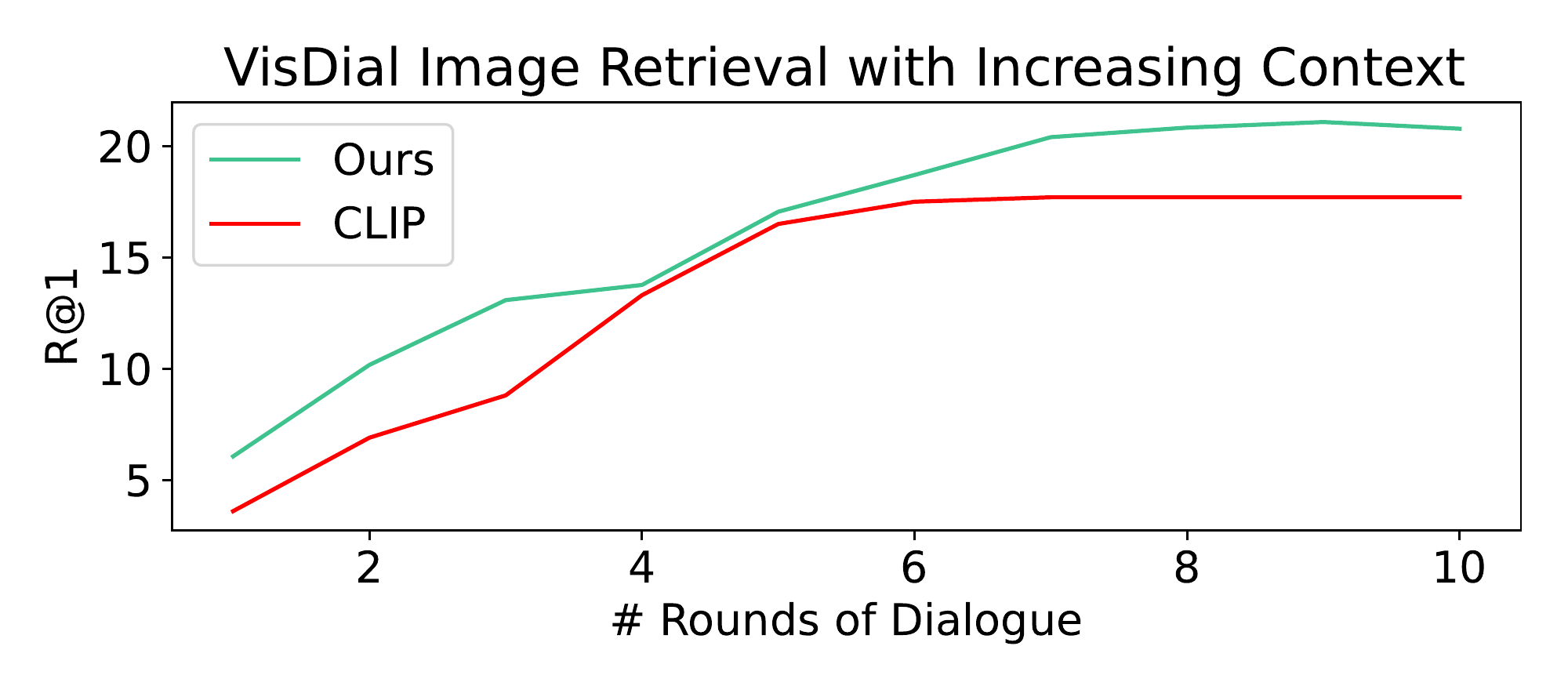}
    \vspace{-0.3in}
    \caption{Increasing input context generally improves performance. Shown are results for image retrieval for VIST~\cite{huang2016visual} (top) and image retrieval on VisDial~\cite{das2017visual} (bottom).
    }
    \label{fig:more_context}
\end{figure}

\paragraph{Multimodal context helps.} Since \ModelName can process interleaved image-text data, a natural question is on the effect of image context compared to text context. To quantify these effects, we run ablations (Fig.~\ref{fig:more_context}, top) varying the number of captions and images provided to the model. We measure the recall of retrieving the correct image conditioned on the provided context for VIST dataset~\cite{huang2016visual}. Increasing context from 1 to 5 captions substantially improves the model: R@1 increases by 5.3\% relative to the single caption case (11.3 to 11.9). However, when we provide an additional image and text example (2 captions + 1 image), we observe an even greater improvement of 30.1\% relative to the single caption case (11.3 to 14.7). This highlights the value of multimodal context: a single image can provide more information than multiple text descriptions.

\paragraph{More context helps.} Performance also steadily improves on image retrieval for VIST as more image and caption context is provided (Fig.~\ref{fig:more_context}, top). The highest R@1 of 18.2 is achieved with 5 captions and 4 images (i.e., the full story context excluding the image to be retrieved), representing a 61.1\% relative improvement over the single caption case. Similar trends are observed for image retrieval using VisDial~\cite{das2017visual} dialogue rounds (Fig.~\ref{fig:more_context}, bottom), with performance improving as more rounds of dialogue are provided. Additionally, the results show that \ModelName outperforms CLIP in all settings, and significantly outperforms CLIP when the full set of dialogue is provided, achieving an improvement of 17.5\% relative over CLIP. These findings suggest that \ModelName is more sensitive to context, enabling it to perform better in situations where correctly parsing long language descriptions is crucial to performance.

\begin{figure}[t]
    \centering
    \includegraphics[width=1.0\linewidth]{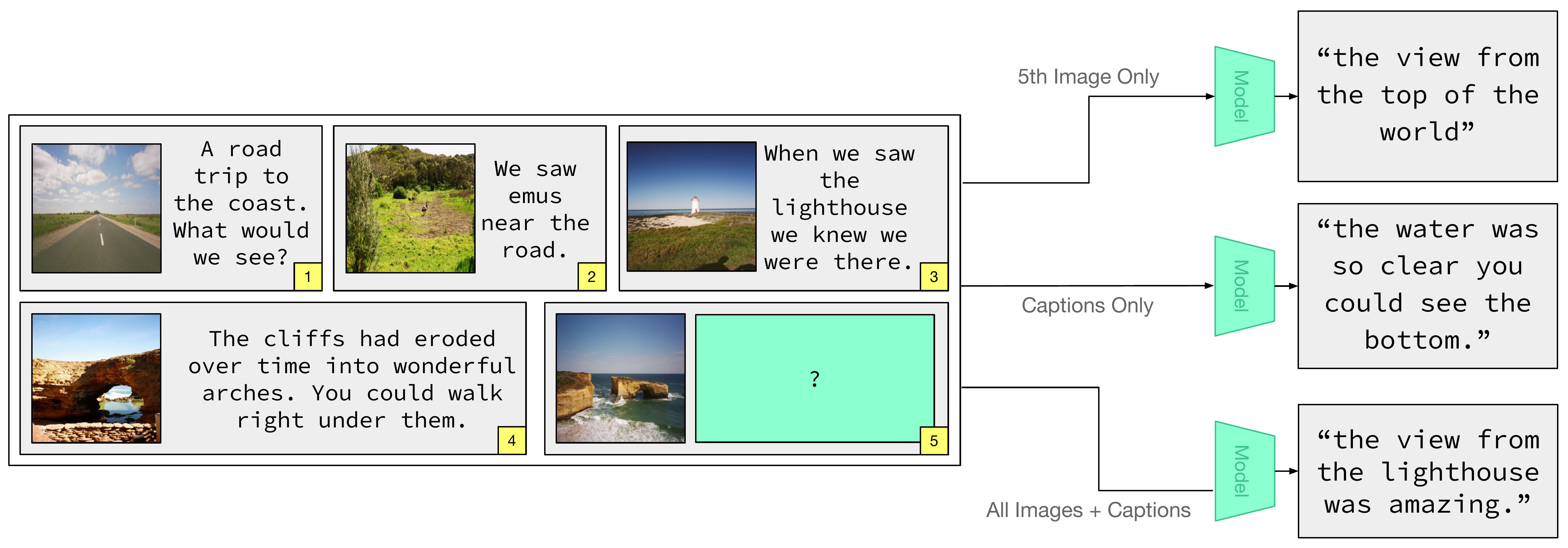}
    \vspace{-0.2in}
    \caption{More coherent and relevant text is generated when in-context examples are provided to \ModelName.
    When multimodal context is provided, the outputs for VIST are more story-like, while the outputs for a single image input are more caption-like.}
    \label{fig:vist_generate_stories}
\end{figure}

\begin{figure}[t]
    \centering
    \includegraphics[width=1.0\linewidth]{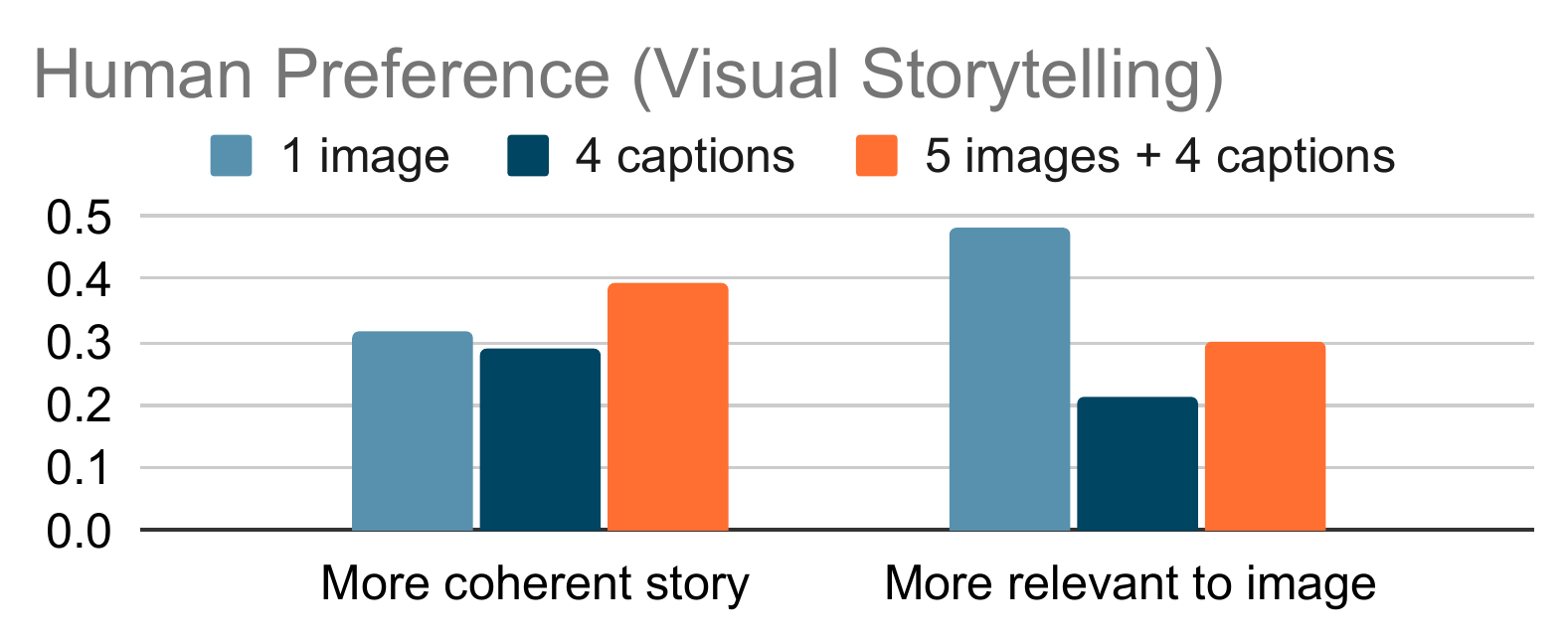}
    \vspace{-0.3in}
    \caption{Human evaluations on VIST story generation. Including both images and captions improves story coherence over using just images, and improves image relevance compared to just captions.}
    \label{fig:human_evals_vist}
\end{figure}

\subsection{In-context Learning and Text Generation}
As \ModelName uses a frozen LLM as its backbone, it is also capable of in-context learning~\cite{brown2020language,chan2022data}, where it generalizes rapidly from a few input examples. We observe this qualitatively from generating new stories for VIST, as shown in Fig.~\ref{fig:vist_generate_stories}. When a single input image is provided, the model generally produces simple caption-like descriptions. However, when prompted with the full multimodal context (i.e., 5 images and 4 stories), the model is able to learn in-context to synthesize plausible story-like text for the 5th image (Fig.~\ref{fig:vist_generate_stories}).

As evaluating generated text is difficult, especially for subjective outputs such as \textit{stories}, we run human evaluations to study the effect of multimodal context on model generated stories. We request human annotators to select the output (from 3 anonymized models) which (1) forms the most coherent story when viewed in relation with the context, and (2) is most relevant to the image. We sample 100 random examples and collect 5 independent ratings each. The results are aggregated from pairwise comparisons (details in appendix) and summarized in Fig.~\ref{fig:human_evals_vist}. When only the last image or the text description is provided as input, the generated stories are rated as less coherent than \ModelName with the full multimodal context (5 images and 4 descriptions). The model generated story is also rated as significantly more relevant to the image inputs compared to the text-only setting, which highlights the ability of the model to condition on both image and text inputs. We observe that the generated output of the single image input case is rated as more relevant compared to the full multimodal context case, which we attribute to the fact that the model produces more factual (albeit less story-like) descriptions (Fig.~\ref{fig:vist_generate_stories}).
These results showcase the ability of \ModelName to learn in-context to synthesize coherent and consistent multimodal outputs.
\section{Future Work}
\ModelName{} is one of the first models capable of parsing image-text inputs, and producing text interleaved with retrieved images. There are several promising directions that are worth exploring in future work. Extending \ModelName{} to perform novel image generation in addition to image retrieval is a natural way to improve its practical capabilities. In our qualitative experiments, we found that the ability of \ModelName{} to produce relevant images was sometimes limited by its candidate retrieval set. This is often the case for prompts that are less likely to occur in natural images, such as fantastical prompts used for benchmarking text-to-image generation models~\cite{yu2022scaling}. On such examples, we find that \ModelName{} (and other retrieval models, such as CLIP) often do not produce relevant images. Developing a model that can both generate text and novel images is an open direction which will likely require further architectural improvements. Another current limitation of \ModelName{} is that it does not always generate \texttt{[RET]} during inference, and generally has a stronger bias to produce regular text tokens. This is likely due to its extensive pretraining on text-only data. During inference, we find that this can be somewhat alleviated by scaling the \texttt{[RET]} logits by a factor $1.3 - 1.5$, prompting with in-context examples, or specifically prompting the model to ask it to show images, which we found helpful in producing good qualitative results. Investigating ways to enable \ModelName to generate \texttt{[RET]} more naturally is also a promising direction for future work. This may entail instruction finetuning~\cite{wei2021finetuned} on multimodal dialogue examples, or training on explicitly interleaved image-text examples~\cite{alayrac2022flamingo}.

\section{Conclusion}

We propose a method to visually ground pretrained frozen language models through efficient finetuning of several linear layers. Our model, \ModelName, is capable of producing coherent interleaved image-text outputs. We show strong zero-shot performance on a variety of tasks involving image-text inputs and outputs, and qualitatively showcase interactive abilities such as multimodal dialogue. These results demonstrate the effectiveness of our approach for bootstrapping general purpose vision-and-language models, capable of consuming and producing arbitrarily interleaved images and text. Scaling \ModelName with larger and more capable language models, training on larger image-text datasets, and extending our approach for generation of novel images from scratch are promising directions for future work.

\section*{Acknowledgements}
This work was partially supported by a gift from Google on Action, Task, and User Journey Modeling,  and supported in part by ONR N000142312368 and DARPA/AFRL FA87502321015. We thank Wendy Kua for help with the figures, and Santiago Cort\'es, Paul Liang, Martin Ma, So Yeon Min, Brandon Trabucco, Saujas Vaduguru, and others for feedback on previous versions of this paper. We thank Felix Hill for insightful discussions about Frozen.

\pagebreak
\bibliography{main}
\bibliographystyle{icml2023}

\pagebreak
\appendix
\section{Qualitative Examples} \label{appendix:qualitative}
\begin{figure*}
\includegraphics[height=0.98\textheight]{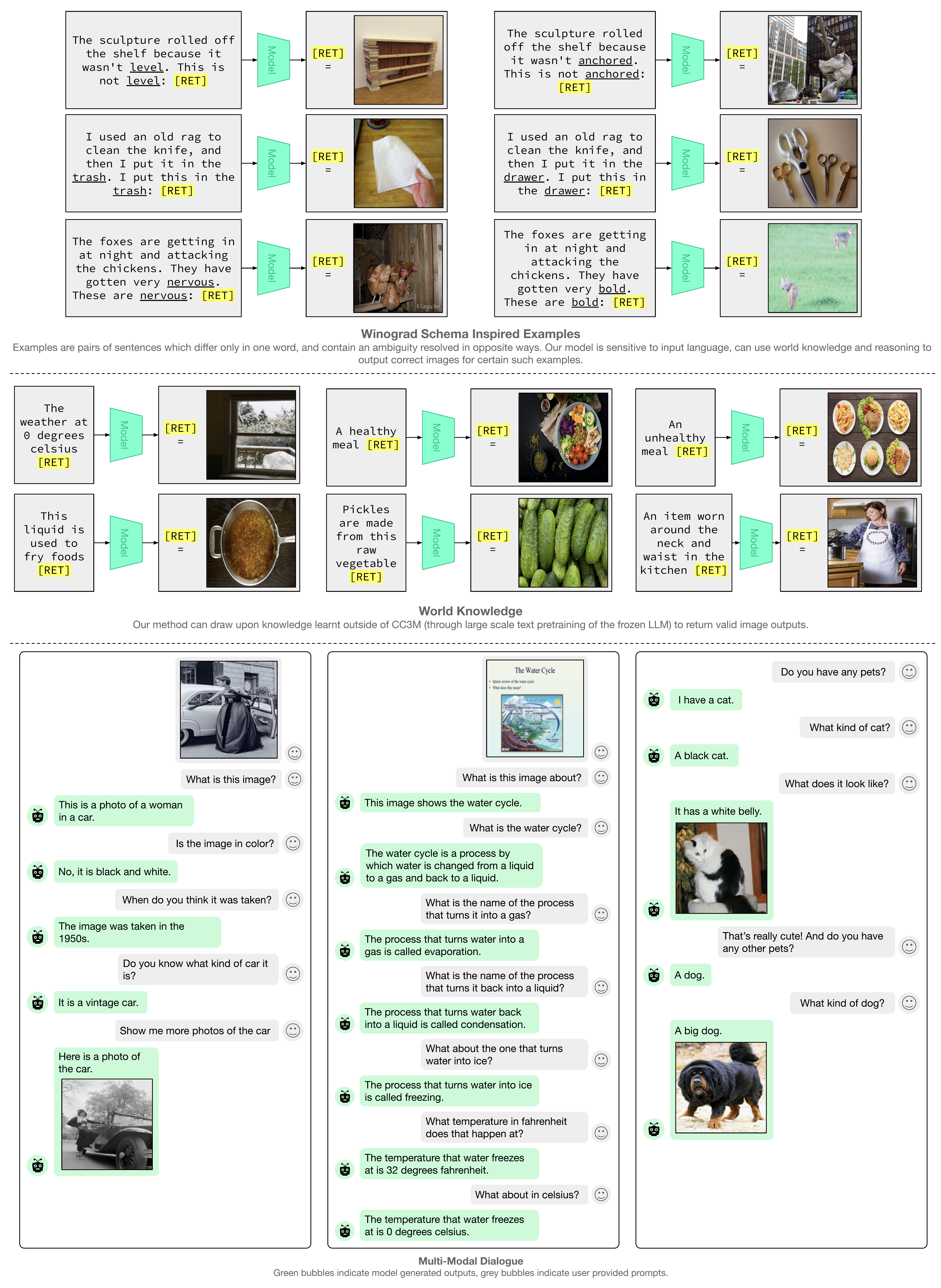}
\vspace{-0.15in}
\caption{Selected examples from \ModelName for various image-text tasks. It is capable of retrieving correct images from some examples from the Winograd schema, as well as possess world knowledge.}
\label{fig:appendix_qualitative_results}
\end{figure*}

In this appendix section, we provide more qualitative examples of \ModelName on various settings.

\paragraph{Sensitivity to prompts.} \ModelName is able to tackle several examples inspired by the Winograd schema~\cite{levesque2012winograd}. These examples contain several sentences which differ only in a single word, and contain an ambiguity resolved in different ways. Our model is capable of retrieving images correctly for different sentences, showcasing its sensitivity to even slight changes in the input prompts.

\paragraph{World knowledge.} The \ModelName approach involves finetuning just linear layers on the Conceptual Captions~\cite{sharma2018conceptual} dataset, which contains image-caption data. Similar to Frozen~\cite{tsimpoukelli2021multimodal}, we find that since our frozen LLM was trained on web-scale text data, it contains knowledge about the world that it can reference for performance on multimodal tasks. For example, we show (Fig.~\ref{fig:appendix_qualitative_results}) that the model knows what the weather at 0 degrees Celsius is likely to look like (snowing), that pickles are made from cucumbers, and more.

\paragraph{Multimodal dialogue.} We also show further examples of our model on dialogue tasks. It is able to reason about input images from the user, as well as respond with semantically appropriate images in the conversation. Similar to its original LLM backbone, it can return coherent text-only outputs. It is able to tap onto its pretrained knowledge to return relevant and accurate information about the world, such as details about the water cycle (second dialogue sequence in Fig.~\ref{fig:appendix_qualitative_results}) and the temperature at which water freezes (in both Fahrenheit and Celsius). This knowledge extends to the visual domain: as seen in the first dialogue sequence in Fig.~\ref{fig:appendix_qualitative_results}, \ModelName is able to understand that the photo is black and white, and likely to be taken in the 1950s.

\subsection{Comparison Against CM3}
\begin{figure*}[t]
\includegraphics[width=1.0\textwidth]{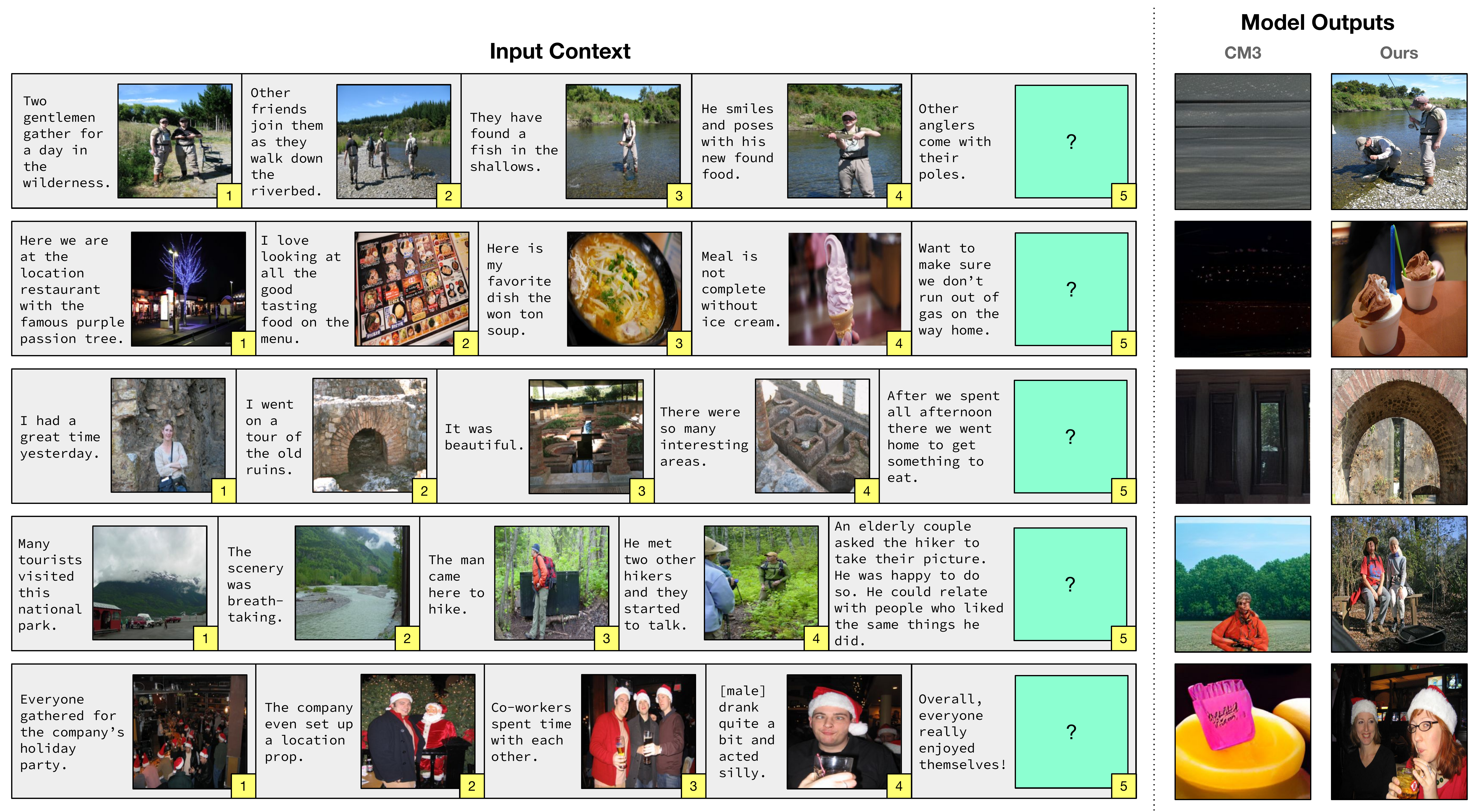}
\vspace{-0.3in}
\caption{Comparison of our model against CM3~\cite{aghajanyan2022cm3} on randomly selected examples from Visual Storytelling (VIST)~\cite{huang2016visual}.}
\label{fig:cm3_vist_qualitative}
\end{figure*}

To the best of our knowledge, CM3~\cite{aghajanyan2022cm3} is the only prior work which proposes a model capable of consuming arbitrarily interleaved image-and-text inputs and generating image-and-text outputs. CM3 trains with far larger computational resources compared to our model -- they train with 384 GPUs for 24 days, while we use a single GPU for 1 day, making our method far more computationally efficient.  

To benchmark the performance of the two models, we run a qualitative comparison to compare the produced images given an image-and-text story input from Visual Storytelling (VIST)~\cite{huang2016visual}. As \ModelName produces images through retrieval and CM3 is a generative model, we are primarily interested in their abilities to produce semantically relevant images, rather than good quality images. Several randomly selected qualitative examples are presented in  Fig.~\ref{fig:cm3_vist_qualitative}. We observe that CM3 is unable to produce coherent outputs for most of the Visual Storytelling inputs. Most outputs produced by CM3 are not interpretable or relevant to the story input. In contrast, the outputs from \ModelName are relevant to the inputs, and a few (e.g., first row of the fishermen, and last row of the people in Santa hats) are capable of retrieving images that are visually and semantically coherent with the input story. We also observed that in general, \ModelName is more sensitive to input prompts and images, while CM3 does not appear to be able to handle long input sequences as well as \ModelName.

\section{Further Analysis}

\subsection{Details on Freezing Ablation}  \label{sec:appendix_frozen_analysis}

We explore the effect of freezing the weights of our language model. Due to GPU memory constraints, we run this experiment with the 1.3b OPT~\cite{zhang2022opt} as the LLM backbone. We compare a version where the weights are kept frozen (\ModelName with a 1.3b LLM), and a version where the language model is allowed to be finetuned. Despite the finetuned model achieving lower loss (on both the training and validation sets of CC3M), we observe that downstream performance on VIST and VisDial significantly deteriorates. On VIST, finetuning the language model decreases retrieval performance on $R@1$ from 12.8 to 6.2, and on VisDial (IT2T), decreases $R@1$ from 14.6 to 1.0. These results demonstrate the importance of freezing the LLM backbone in order to retain the abilities of the LLM (in-context learning, zero-shot generalization) learnt from large scale text pretraining.

\subsection{Joint Retrieval + Captioning Training}  \label{sec:retrieval_captioning_analysis}

\begin{table*}[t]
\begin{center}
\setlength{\tabcolsep}{3pt}
\scriptsize
\resizebox{1.0\linewidth}{!}{%
\begin{tabular}{lccccccccccccc}
\noalign{\smallskip}
\toprule
 & \multicolumn{5}{c}{\textbf{Captioning}} &  & \multicolumn{3}{c}{\textbf{T2I Retrieval}} & & \multicolumn{3}{c}{\textbf{I2T Retrieval}} \\
\cmidrule{2-6} \cmidrule{8-10} \cmidrule{12-14}
\textbf{Training Loss} & BLEU-1 & BLEU-2 & BLEU-3 & BLEU-4 & METEOR   &  &  R@1  & R@5  & R@10 &  &  R@1  & R@5 & R@10 \\ \midrule
 Captioning &   0.4768  &  0.2899  &   0.1664 &   0.0965  &  0.2820  & &  -  &   - & - & & -  &  - & - \\
 Retrieval &   -  &  -  &  - &   -  &  -  & &  23.4  &  47.3 &  59.0  & &  26.8  &  52.4 & 63.6 \\
 Captioning + Retrieval &   0.4766  &  0.2932  &   0.1720 &   0.1023  &  0.2873  & &  23.4   &  47.2 & 58.0 & & 26.4  &   52.3 &  63.4 \\
\bottomrule
\end{tabular}
}
\vspace{-0.1in}
\caption{Ablation results over different training objectives. All models are trained on CC3M~\cite{sharma2018conceptual} and reported on the 5K validation set of MS-COCO (2017)~\cite{lin2014microsoft}. For captioning, we report BLEU~\cite{papineni2002bleu} and METEOR~\cite{banerjee2005meteor} scores, and for retrieval, we report Recall$@k$ (single captions).}
\label{tab:training_obj_ablation}
\end{center}
\end{table*}

We run ablations over the different loss functions (Tab.~\ref{tab:training_obj_ablation}). As the retrieval only model is only able to process text inputs, and the captioning model is only able to generate text outputs, we are unable to test the ablated models on VIST or VisDial. Hence, we benchmark their captioning and retrieval performance on MS-COCO~\cite{lin2014microsoft}, which tests generalization from our training data (Conceptual Captions 3M~\cite{sharma2018conceptual}).

We find that joint training with the multi-task captioning and retrieval losses have no negative effect on performance on the individual tasks, with most metrics staying the same (with captioning scores slightly improving in the \ModelName model), which shows that we do not suffer from optimizing our model over multiple objectives.

\subsection{Image-Text Concatenation for Captioning}  \label{sec:concat_analysis}

During training, we concatenate distinct examples sequentially for image captioning. We found that this was significantly helpful for several downstream tasks, as it encourages our model to attend to multiple images within a sequence during training. In particular, on the VIST dataset, enabling image-text concatenation improves $R@1$ from 11.6 to 15.6 when 5 captions and 4 images are provided as input (see Sec.~\ref{sec:contextual_image_retrieval}). On VisDial, we find that the ablated model performs similarly. This agrees with intuition, as VIST requires processing of multiple images interleaved with text (while VisDial only has a single image in its input).

These results show that random concatenation is a strong data augmentation strategy for generalization to tasks involving interleaved-image-text data. As large datasets with interleaved image-text data (such as those used in Flamingo~\cite{alayrac2022flamingo} or CM3~\cite{aghajanyan2022cm3}) are generally not available to the public, our proposed approach may be a way to leverage large open image-text datasets~\cite{schuhmann2021laion} for training such multimodal models.

\subsection{Image-Text Concatenation for Retrieval}  \label{sec:appendix_concat_ret_analysis}

As described in Sec.~\ref{sec:concat_analysis}, we concatenate distinct examples sequentially for image captioning during training. This was found to be helpful in encouraging the model to learn to attend to multiple images interleaved within an image-text sequence. We ran the same experiment for concatenating examples for both image captioning and image-text retrieval. For retrieval on concatenated examples, the model is tasked to retrieve two separate images, one for the \texttt{[RET]} token at the end of the first caption, and one for the one at the end of the second. An example input text for a concatenated example is:
\begin{verbatim}
silhouette of a plane against the 
sunset [RET] cute cat sitting on a
scooter [RET]
\end{verbatim}
in which case the model is expected to retrieve the appropriate images of a plane and a cat respectively. However, we find that this concatenation does not appear to have a positive effect on the downstream tasks of VIST and VisDial. On VIST, $R@1$ also slightly decreases from 15.6 to 14.4. We hypothesize that this is likely because these tasks (and many of our qualitative examples) do not require the model to retrieve \textit{disparate} images -- multiple image outputs are usually related (e.g., dog example in Fig.~\ref{fig:qualitative_results} of the main paper) and involve coreferencing. Concatenation of retrieval examples during training is likely to deteriorate these abilities. However, in certain multimodal applications (such as retrieval for more factual multimodal tasks, rather than dialogue), it may be possible that this retrieval concatenation strategy is still useful, and we leave exploration to future work.


\subsection{Scaling Properties}  \label{sec:scaling}
\begin{figure}
    \centering
    \includegraphics[width=0.9\linewidth]{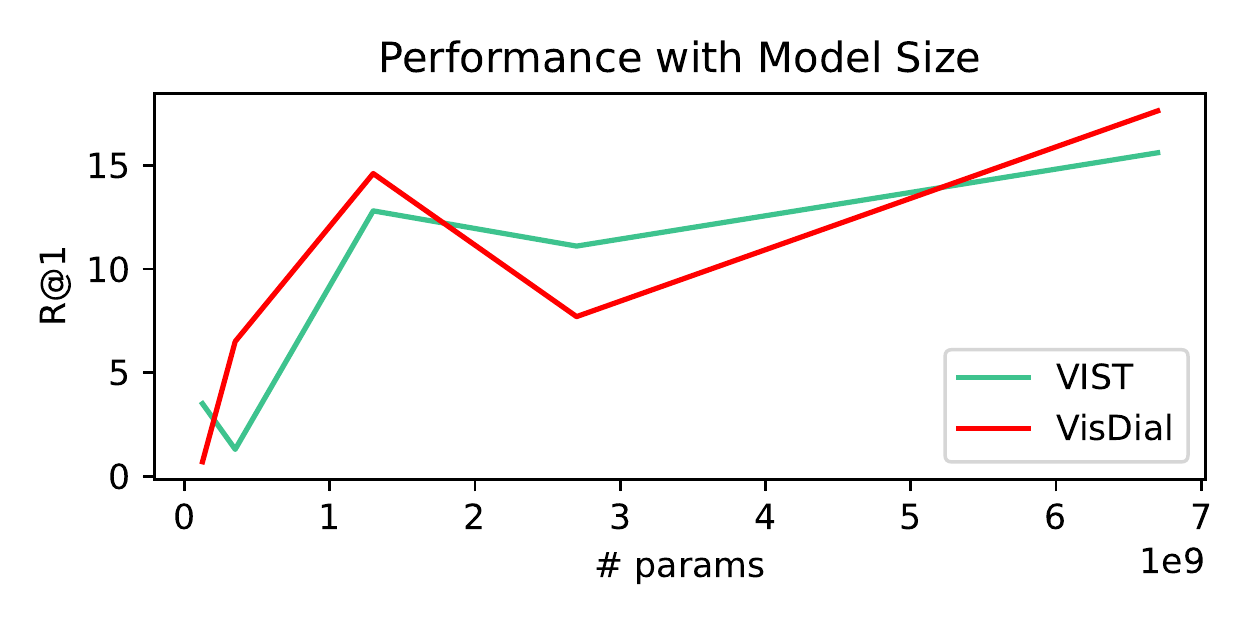}
    \caption{Performance on VIST contextual image retrieval and VisDial IT2T over different model scales. Performance generally improves as models get bigger.}
    \label{fig:scaling_curve}
\end{figure}

\ModelName is trained in a model-agnostic approach (Sec.~\ref{sec:training_setup}), which can be applied to any pre-trained text-only LLM. We demonstrate that our approach is scalable and can benefit from larger, more expressive LLMs. We conduct experiments using the OPT model family~\cite{zhang2022opt} with models of increasing parameter counts (125M, 350M, 1.3B, 2.7B, and 6.7B).

Our results, as shown in Fig.~\ref{fig:scaling_curve}, indicate that performance generally improves with increasing model size on the zero-shot contextual image retrieval task for Visual Storytelling~\cite{huang2016visual} and Visual Dialog~\cite{das2017visual}. This promising trend suggests that our framework is likely to benefit from even larger text-only models such as GPT-3 (175B)~\cite{brown2020language}, Chinchilla (70B)~\cite{hoffmann2022training}, or PaLM (540B)~\cite{chowdhery2022palm}. In future work, it will be interesting to test this, and trained models may learn additional interesting emergent behaviors~\cite{wei2022emergent}.

\subsection{Text Generation Results}

In addition to the above evaluations, we also ran the zero-shot VQAv2~\cite{goyal2017making}. We apply the same normalization techniques from their GitHub repo\footnote{\url{https://github.com/GT-Vision-Lab/VQA}}, and prompt the model with the prefix \texttt{Q: \{question\} A:} format. On zero-shot VQA, our model achieves a score of 28.51, which is better or comparable to prior methods: a reimplementation of Frozen achieves 25.53, while running the MAGMA pretrained model (which uses 25M image-text as training data, including the VQA training set), achieves 28.35. These results show that our approach is competitive with similar methods that use parameter efficient adaptation (Frozen, MAGMA), with the added bonus that our model can perform image retrieval interleaved with text. This allows us to handle a much wider range of tasks: for example, Frozen and MAGMA cannot produce images interleaved with generated text. Our model is also significantly more efficient (1 GPU day of training) compared to prior work (MAGMA uses 32 GPUs for 1.25 days).

\section{Human Evaluation Procedure}
\begin{figure*}
    \centering
    \includegraphics[width=0.9\textwidth]{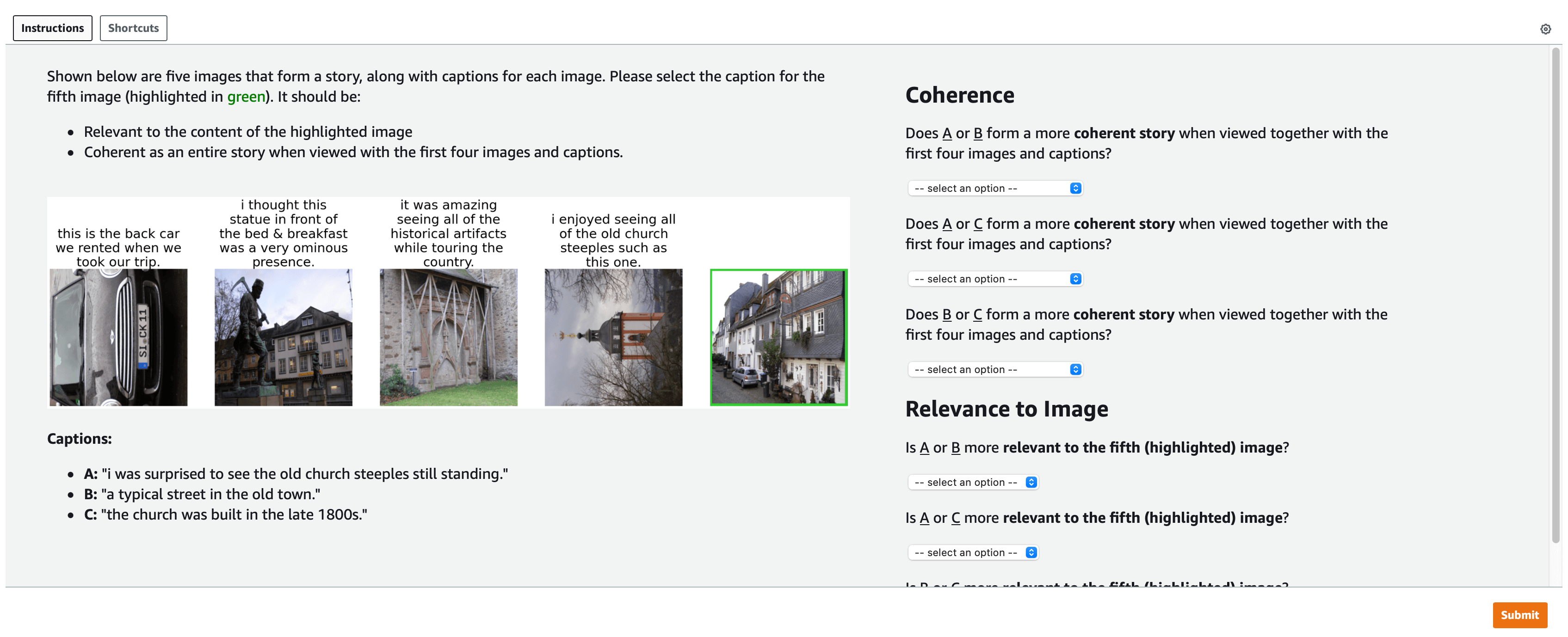}
    \vspace{-0.15in}
    \caption{User interface shown to for human raters for performing evaluations. Raters are tasked to compare model outputs with different contexts as inputs, and rate (1) whether they form more coherent stories and (2) are more relevant to the last image. Model outputs are anonymized and shuffled.}
    \label{fig:human_evals_ui}
\end{figure*}

\begin{figure*}
    \centering
    \includegraphics[width=0.33\textwidth]{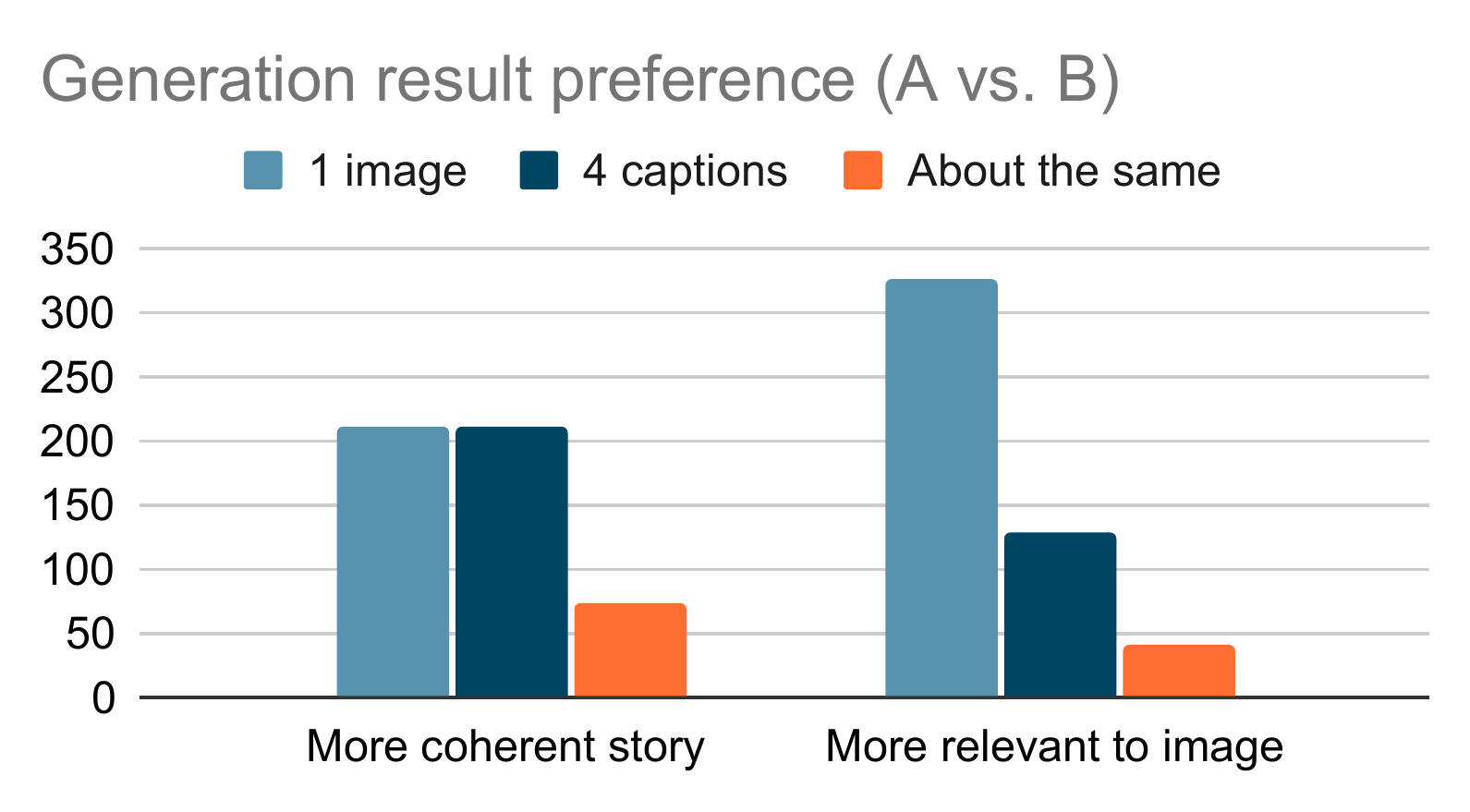}
    \includegraphics[width=0.33\textwidth]{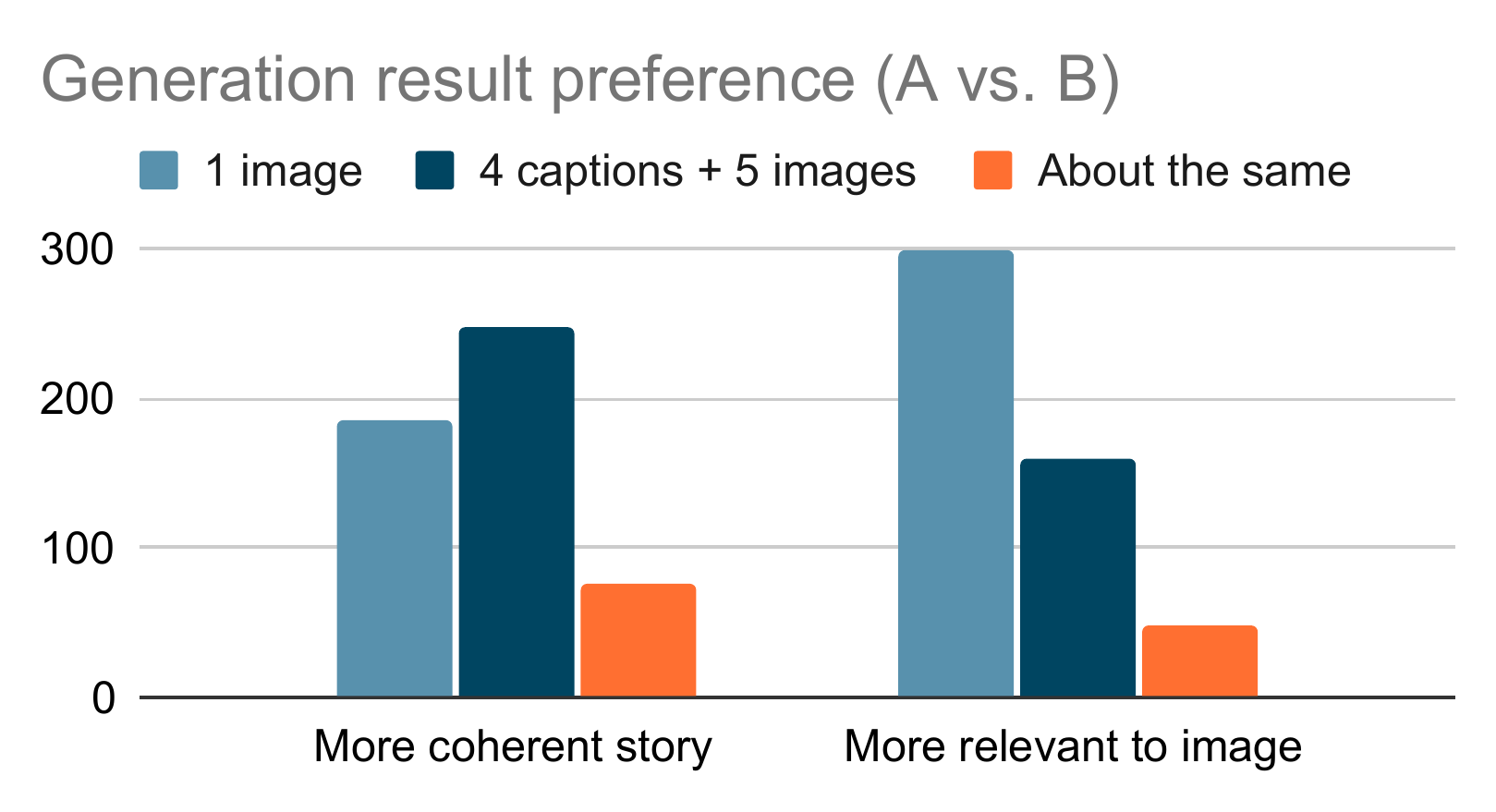}
    \includegraphics[width=0.33\textwidth]{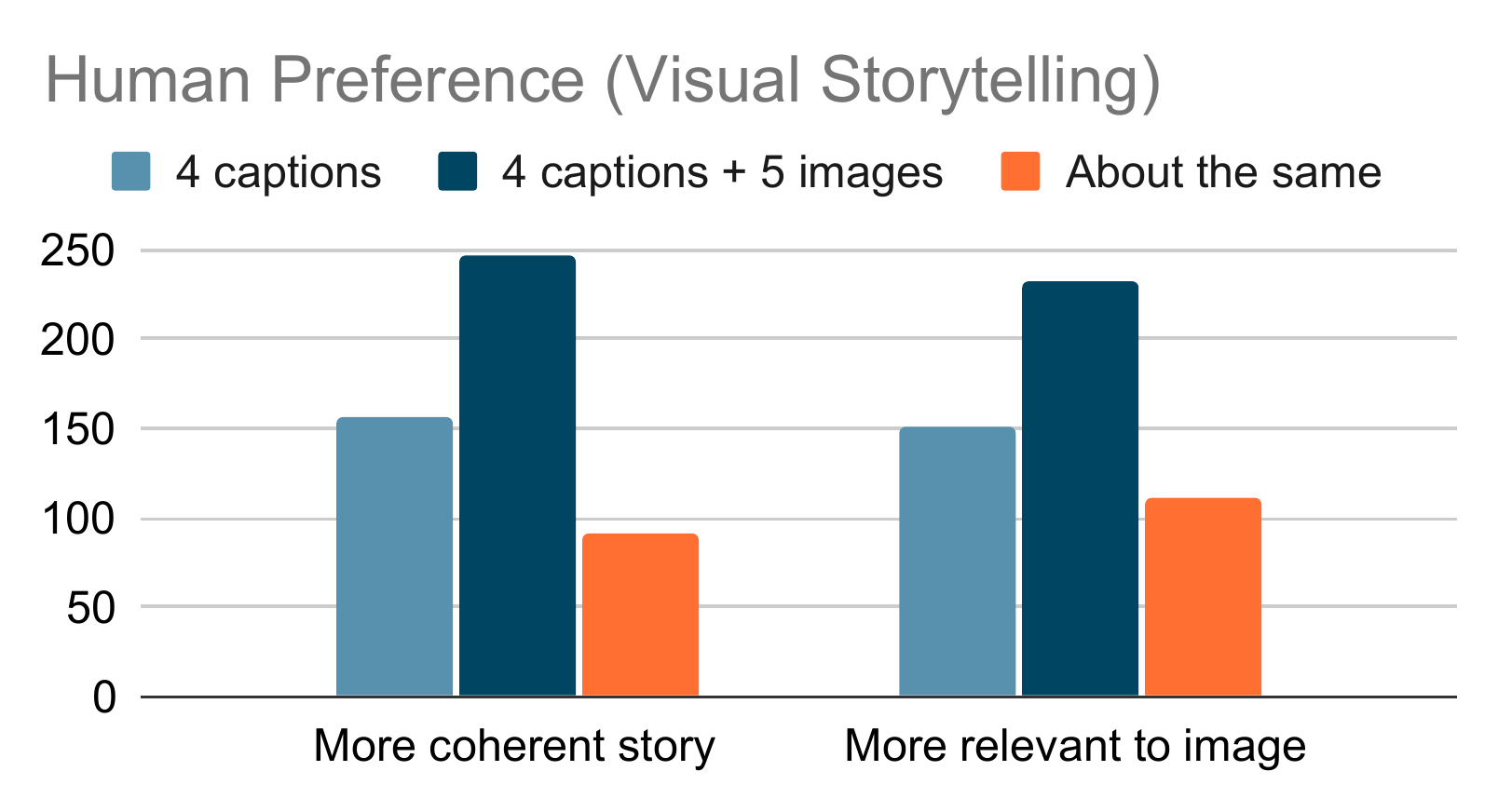}
    \vspace{-0.2in}
    \caption{Head-to-head evaluations of \ModelName with different input contexts. For each figure, human evaluators are tasked to select whether one model is more coherent than the other, and if one is more relevant to the image.}
    \label{fig:human_evals_individual}
\end{figure*}

\begin{figure*}
    \centering
    \includegraphics[width=1.0\textwidth,height=3.8in]{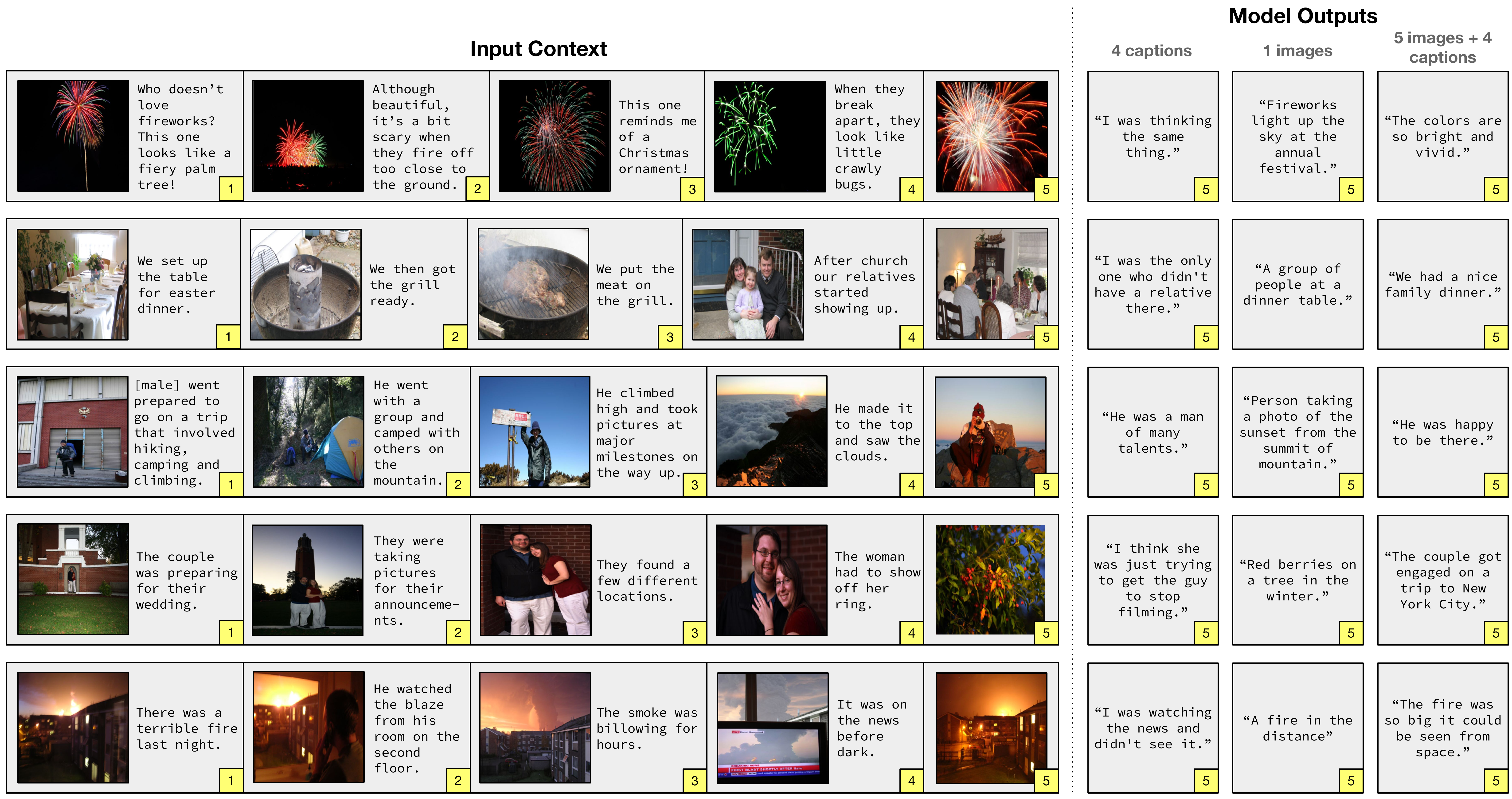}
    \vspace{-0.2in}
    \caption{Examples of generated stories conditioned on different input contexts.}
    \label{fig:vist_qualitative}
\end{figure*}

As detailed in Sec.~\ref{sec:multimodal_context} of the main paper, we perform human evaluations of 500 randomly selected examples to determine generated story quality given different input contexts. Users are tasked to evaluate three settings:
\begin{enumerate}
    \setlength\itemsep{0em}
    \item Generated outputs conditioned on the last image only.
    \item Generated outputs conditioned on the preceding text descriptions only.
    \item Generated outputs conditioned on the preceding images and text descriptions.
\end{enumerate}
The results (described in Sec.~\ref{sec:multimodal_context} of the main paper), show that setting \#3, which contains the full multimodal context, produces text that is rated as more coherent than both settings \#1 and \#2, which contain less context (which is of a single modality). We present individual ratings in Fig.~\ref{fig:human_evals_individual}, which summarize the head-to-head comparisons for comparing one model against another.

We observe that \#3 produces descriptions that are rated as more relevant to the image than \#2, but less relevant than the single image case \#1. We attribute this to \#1 generating more factual, caption-like outputs (as only a single image is provided as input), while \#3 generates more story-like outputs (and hence are more coherent overall). Overall, these results show the ability of \ModelName to learn in-context to produce stories rather than caption outputs. \ModelName is able to learn in-context and condition on both the input images and text descriptions to produce coherent story outputs which are aligned to a corresponding image (side-by-side qualitative examples in Fig.~\ref{fig:vist_qualitative}).

We ran evaluations on Amazon Mechanical Turk with human raters located in the US and Canada. Annotators were paid at an estimated hourly rate of 15 USD / hr. We spent a total of approximately 140 USD to collect our evaluations.

\section{Current Limitations and Broader Impacts}  \label{appendix:failure_modes_analysis}

Many existing large language models and large generative models are prone to certain types of unintended behavior. They sometimes make up facts, generate toxic and socially biased text outputs, propagate disinformation, or ignore user prompts~\cite{gehman2020realtoxicityprompts,bommasani2021opportunities,bender2021dangers}. When tasked to generate text, these large models also often exhibit failure modes such as neural text degeneration and generation of incoherent and repetitive text~\cite{holtzman2019curious,li2022contrastive}.

A core component of the \ModelName model is the frozen LLM backbone, which we ground for producing text and visual outputs. Unsurprisingly, our model also inherits some of the problems that text-only LLMs possess, such as generating repetitive outputs, not following user instructions, and other common failure modes. It is also susceptible to the limitations of these language models, including the risk of producing disinformation or toxic content. The broader issues relating to text generation for our model are also likely to be addressed and alleviated by future work on better large language models. In particular, using language models that are finetuned with human feedback~\cite{ouyang2022training} or instruction finetuned~\cite{wei2021finetuned,chung2022scaling} may be one direction towards improving output quality, and for reducing the risk of producing toxic and socially biased content. As \ModelName is modular in nature, we can easily swap out our LLM backbone for better and more robust language models released in the future, enabling us to easily improve its performance on downstream applications, and reduce the risk of generating harmful content.

\ModelName is also a model that can produce images. In this work, we produce images (interleaved within text) by retrieving from a fixed set of images from Conceptual Captions~\cite{sharma2018conceptual}. Like other image-text retrieval models~\cite{radford2021learning,jia2021scaling}, our model is susceptible to existing biases found in the training and retrieval datasets. Although image retrieval is unable to produce truly novel images from outside of the retrieval data, it also has benefits for controlling output results. Unlike image generation models which synthesize novel images from scratch, a benefit of retrieving from a fixed corpus is that it allows us to explicitly control what our model can output. Retrieval enables possible mitigation strategies such as filtering for inappropriate content, such that \ModelName and similar models would not be able to produce particular types of objectionable images. However, for deployment of such technologies (and future research on generative multimodal dialogue models), it is essential to test and analyze data used to mitigate the risk of training large multimodal models~\cite{birhane2021multimodal}. This will involve filtering of images, rigorous testing of model biases (for both image and text content), and more.




\end{document}